\documentclass{article} % For LaTeX2e
\usepackage[utf8]{inputenc}
\usepackage{iclr2017_conference,times}
\usepackage{hyperref}
\usepackage{url}
\usepackage{bm}
\usepackage{graphicx}
\usepackage[cmex10]{amsmath}
\usepackage{amssymb,amsthm}
\usepackage[subrefformat=parens]{subcaption}
\usepackage[compact]{titlesec}

\usepackage[backend=biber,style=authoryear,natbib=true,mincrossrefs=1000]{biblatex}
\AtBeginDocument{\toggletrue{blx@useprefix}}
\AtBeginBibliography{\togglefalse{blx@useprefix}}
\usepackage{tikz}
\usetikzlibrary{backgrounds,calc,chains,fit,matrix,positioning,shadows,shapes.misc,circuits.ee}
\tikzset{input/.style={}}
\tikzset{output/.style={}}
\tikzset{operator/.style={circle, draw, minimum size=2.5ex, inner sep=0pt}}
\tikzset{filter/.style={rectangle, draw, fill=white, minimum size=3.5ex, inner xsep=1.5ex}}
\tikzset{other/.style={rounded rectangle, draw, fill=white, minimum size=3.5ex, inner xsep=1ex}}
\tikzset{branch/.style={circle, draw, fill=black, thick, minimum size=.5ex, inner sep=0pt}}
\tikzset{rv/.style={circle, draw, thick, fill=white, minimum size=2.75ex, inner sep=0pt}}
\tikzset{ob/.style={circle, draw, thick, fill=lightgray, minimum size=2.75ex, inner sep=0pt}}
\tikzset{pa/.style={circle, draw, thick, fill=black, minimum size=1ex, inner sep=0pt}}
\tikzset{/tikz/thin/.style={line width=.6pt}}
\tikzset{/tikz/thick/.style={line width=1pt}}
\tikzset{every path/.style={thin}}
\tikzset{>=direction ee}

\title{End-to-end Optimized Image Compression}

\author{Johannes Ballé\thanks{JB and EPS are supported by the Howard Hughes Medical Institute.} \\
Center for Neural Science \\
New York University \\
New York, NY 10003, USA \\
\texttt{johannes.balle@nyu.edu} \\
\And
Valero Laparra \\
Image Processing Laboratory \\
Universitat de València \\
46980 Paterna, Spain \\
\texttt{valero.laparra@uv.es} \\
\And
Eero P. Simoncelli\footnotemark[1] \\
Center for Neural Science and Courant Institute of Mathematical Sciences \\
New York University \\
New York, NY 10003, USA \\
\texttt{eero.simoncelli@nyu.edu}
}

\DeclareMathOperator{\round}{round}

\newcommand{\E}{\operatorname{\mathbb E}}
\newcommand{\R}{\operatorname{\mathbb R}}
\newcommand{\Z}{\operatorname{\mathbb Z}}
\newcommand{\D}{\;\mathrm{d}}

\newcommand{\const}{\mathrm{const}}

\allowdisplaybreaks[2]
\graphicspath{{figures/}}

\iclrfinalcopy % Uncomment for camera-ready version

\begin{document}
\suppressfloats % no floats on title page

\maketitle

\begin{abstract}
We describe an image compression method, consisting of a nonlinear analysis transformation, a uniform quantizer, and a nonlinear synthesis transformation. The transforms are constructed in three successive stages of convolutional linear filters and nonlinear activation functions. Unlike most convolutional neural networks, the joint nonlinearity is chosen to implement a form of local gain control, inspired by those used to model biological neurons.
Using a variant of stochastic gradient descent, we jointly optimize the entire model for rate--distortion performance over a database of training images, introducing a continuous proxy for the discontinuous loss function arising from the quantizer.
Under certain conditions, the relaxed loss function may be interpreted as the log likelihood of a generative model, as implemented by a variational autoencoder. Unlike these models, however, the compression model must operate at any given point along the rate--distortion curve, as specified by a trade-off parameter.
Across an independent set of test images, we find that the optimized method generally exhibits better rate--distortion performance than the standard JPEG and JPEG~2000 compression methods. More importantly, we observe a dramatic improvement in visual quality for all images at all bit rates, which is supported by objective quality estimates using MS-SSIM.
\end{abstract}

%%% ==============================================
\section{Introduction}
Data compression is a fundamental and well-studied problem in engineering, and is commonly formulated with the goal of designing codes for a given discrete data ensemble with minimal entropy~\citep{Sh48}. The solution relies heavily on knowledge of the probabilistic structure of the data, and thus the problem is closely related to probabilistic source modeling. However, since all practical codes must have finite entropy, continuous-valued data (such as vectors of image pixel intensities) must be quantized to a finite set of discrete values, which introduces error. In this context, known as the \emph{lossy compression problem}, one must trade off two competing costs: the entropy of the discretized representation (\emph{rate}) and the error arising from the quantization (\emph{distortion}). Different compression applications, such as data storage or transmission over limited-capacity channels, demand different rate--distortion trade-offs.

Joint optimization of rate and distortion is difficult.
Without further constraints, the general problem of optimal quantization in high-dimensional spaces is intractable~\citep{GeGr92}. For this reason, most existing image compression methods operate by linearly transforming the data vector into a suitable continuous-valued representation, quantizing its elements independently, and then encoding the resulting discrete representation using a lossless \emph{entropy code}~\citep{Wi72,NeLi80}. This scheme is called \emph{transform coding} due to the central role of the transformation. For example, JPEG uses a discrete cosine transform on blocks of pixels, and JPEG~2000 uses a multi-scale orthogonal wavelet decomposition. Typically, the three components of transform coding methods -- transform, quantizer, and entropy code -- are separately optimized (often through manual parameter adjustment).

\begin{figure}
\centering%
\begin{tikzpicture}[x=3.5cm,y=1.2cm]

\node[branch,minimum size=4pt] (x) {};
\node[above left=.3ex of x,text height=1.5ex,text depth=.25ex] (lbl_x) {\large $\bm x$};
\node[branch,minimum size=4pt] (x_hat) at ($(x)+(0,-1)$) {};
\node[above left=.3ex of x_hat,text height=1.5ex,text depth=.25ex] {\large $\bm{\hat x}$};
\node[branch,minimum size=4pt] (y) at ($(x)+(1,0)$) {};
\node[above left=.3ex of y,text height=1.5ex,text depth=.25ex] (lbl_y) {\large $\bm y$};
\node[branch,minimum size=4pt] (y_hat) at ($(y)+(0,-1)$) {};
\node[above left=.3ex of y_hat,text height=1.5ex,text depth=.25ex] {\large $\bm{\hat y}$};
\node[branch,minimum size=4pt] (q) at ($(y)+(.5,-.5)$) {};
\node[above right=.3ex of q] {\large $\bm q$};
\node[branch,minimum size=4pt] (z) at ($(x)+(-1,0)$) {};
\node[above left=.3ex of z,text height=1.5ex,text depth=.25ex] (lbl_z) {\large $\bm z$};
\node[branch,minimum size=4pt] (z_hat) at ($(z)+(0,-1)$) {};
\node[above left=.3ex of z_hat,text height=1.5ex,text depth=.25ex] {\large $\bm{\hat z}$};

\path (x) -- (y) node[filter,midway] (g_a) {\large $g_a$};
\path (y_hat) -- (x_hat) node[filter,midway] (g_s) {\large $g_s$};
\path (x) -- (z) node[filter,midway] (g_p1) {\large $g_p$};
\path (x_hat) -- (z_hat) node[filter,midway] (g_p2) {\large $g_p$};

\draw[shorten <=1ex] (x) -- (g_a);
\draw[->,shorten >=1ex] (g_a) -- (y);
\draw[->,shorten <=1ex,shorten >=1ex,rounded corners=1.5ex] (y) -- (q|-y) -- (q);
\draw[->,shorten <=1ex,shorten >=1ex,rounded corners=1.5ex] (q) -- (q|-y_hat) -- (y_hat);
\draw[shorten <=1ex] (y_hat) -- (g_s);
\draw[->,shorten >=1ex] (g_s) -- (x_hat);
\draw[shorten <=1ex] (x) -- (g_p1);
\draw[->,shorten >=1ex] (g_p1) -- (z);
\draw[shorten <=1ex] (x_hat) -- (g_p2);
\draw[->,shorten >=1ex] (g_p2) -- (z_hat);

\node[text width=11ex,align=center,above] (lbl_d) at ($(x_hat)+(0,-1)$) {data \\ space};
\node[text width=11ex,align=center,above] (lbl_c) at ($(y_hat)+(0,-1)$) {code \\ space};
\node[text width=11ex,align=center,above] (lbl_p) at ($(z_hat)+(0,-1)$) {perceptual \\ space};

\node (R) at ($(lbl_c)+(.75,0)$) {\large $R$};
\draw[<-,densely dotted,thick,shorten >=1ex,rounded corners=1.5ex] (R) -- (R|-q) -- (q);
\node (D) at ($(lbl_p)+(-.5,0)$) {\large $D$};
\draw[<-,densely dotted,thick,shorten >=1ex,rounded corners=1.5ex] (D) -- (D|-z) -- (z);
\draw[<-,densely dotted,thick,shorten >=1ex,rounded corners=1.5ex] (D) -- (D|-z_hat) -- (z_hat);

\begin{pgfonlayer}{background}
\node[rounded corners=2ex,fill=black,opacity=.1,fit=(lbl_x)(lbl_d),inner xsep=0pt,inner ysep=1ex] (data_space) {};
\node[rounded corners=2ex,fill=black,opacity=.1,fit=(lbl_y)(lbl_c),inner xsep=0pt,inner ysep=1ex] (code_space) {};
\node[rounded corners=2ex,fill=black,opacity=.1,fit=(lbl_z)(lbl_p),inner xsep=0pt,inner ysep=1ex] (perceptual_space) {};
\end{pgfonlayer}
\end{tikzpicture}
\caption{General nonlinear transform coding framework~\citep{BaLaSi16}.
A vector of image intensities $\bm x \in \R^N$ is mapped to a latent \emph{code space} via a parametric \emph{analysis} transform, $\bm y =g_a(\bm x; \bm\phi)$. This representation is quantized, yielding a discrete-valued vector $\bm q \in \Z^M$ which is then compressed. The \emph{rate} of this discrete code, $R$, is lower-bounded by the entropy of the discrete probability distribution of the quantized vector, $H[P_{\bm q}]$. To reconstruct the compressed image, the discrete elements of $\bm q$ are reinterpreted as a continuous-valued vector $\bm{\hat y}$, which is transformed back to the data space using a parametric \emph{synthesis} transform $\bm{\hat x} = g_s(\bm{\hat y}; \bm\theta)$. Distortion is assessed by transforming to a \emph{perceptual space} using a (fixed) transform, $\bm{\hat z} = g_p(\bm{ \hat x})$, and evaluating a metric $d(\bm z,\bm{\hat z})$. We optimize the parameter vectors $\bm \phi$ and $\bm \theta$ for a weighted sum of the rate and distortion measures, $R+\lambda D$, over a set of images.}%
\label{fig:overview}
\end{figure}

We have developed a framework for end-to-end optimization of an image compression model based on \emph{nonlinear} transforms (figure~\ref{fig:overview}). Previously, we demonstrated that a model consisting of linear--nonlinear block transformations, optimized for a measure of perceptual distortion, exhibited visually superior performance compared to a model optimized for mean squared error (MSE)~\citep{BaLaSi16}. Here, we optimize for MSE, but use a more flexible transforms built from cascades of linear convolutions and nonlinearities. Specifically, we use a generalized divisive normalization (GDN) joint nonlinearity that is inspired by models of neurons in biological visual systems, and has proven effective in Gaussianizing image densities \citep{BaLaSi15}. This cascaded transformation is followed by uniform scalar quantization (i.e., each element is rounded to the nearest integer), which effectively implements a parametric form of vector quantization on the original image space. The compressed image is reconstructed from these quantized values using an approximate parametric nonlinear inverse transform.

For any desired point along the rate--distortion curve, the parameters of both analysis and synthesis transforms are jointly optimized using stochastic gradient descent. To achieve this in the presence of quantization (which produces zero gradients almost everywhere), we use a proxy loss function based on a continuous relaxation of the probability model, replacing the quantization step with additive uniform noise. The relaxed rate--distortion optimization problem bears some resemblance to those used to fit generative image models, and in particular variational autoencoders~\citep{KiWe14,ReMoWi14}, but differs in the constraints we impose to ensure that it approximates the discrete problem all along the rate--distortion curve. Finally, rather than reporting differential or discrete entropy estimates, we implement an entropy code and report performance using actual bit rates, thus demonstrating the feasibility of our solution as a complete lossy compression method.

%%% ==============================================
\section{Choice of forward, inverse, and perceptual transforms}
Most compression methods are based on orthogonal linear transforms, chosen to reduce correlations in the data, and thus to simplify entropy coding. But the joint statistics of linear filter responses exhibit strong higher order dependencies. These may be significantly reduced through the use of joint local nonlinear gain control operations \citep{ScSi01,Ly10,SiBe13}, inspired by models of visual neurons \citep{He92,CaHe12}. Cascaded versions of such models have been used to capture multiple stages of visual transformation \citep{SiHe98,MaBoCa08}. Some earlier results suggest that incorporating local normalization in linear block transform coding methods can improve coding performance~\citep{MaEpNaSi06}, and can improve object recognition performance of cascaded convolutional neural networks~\citep{JaKaRaLe09}. However, the normalization parameters in these cases were not optimized for the task. Here, we make use of a generalized divisive normalization (GDN) transform with optimized parameters, that we have previously shown to be highly efficient in Gaussianizing the local joint statistics of natural images, much more so than cascades of linear transforms followed by pointwise nonlinearities~\citep{BaLaSi15}.

Note that some training algorithms for deep convolutional networks incorporate ``batch normalization'', rescaling the responses of linear filters in the network so as to keep it in a reasonable operating range~\citep{IoSz15}. This type of normalization is different from local gain control in that the rescaling factor is identical across all spatial locations. Moreover, once the training is completed, the scaling parameters are typically fixed, which turns the normalization into an affine transformation with respect to the data -- unlike GDN, which is spatially adaptive and can be highly nonlinear.

Specifically, our analysis transform $g_a$ consists of three stages of convolution, subsampling, and divisive normalization. We represent the $i$th input channel of the $k$th stage at spatial location $(m,n)$ as $u_i^{(k)}(m,n)$. The input image vector $\bm x$ corresponds to $u_i^{(0)}(m,n)$, and the output vector $\bm y$ is $u_i^{(3)}(m,n)$. Each stage then begins with an affine convolution:
\begin{align}
\label{eq:conv}
v_i^{(k)}(m, n) &= \sum_j \bigl(h_{k,ij} \ast u_j^{(k)}\bigr)(m, n) + c_{k,i},
\intertext{where $\ast$ denotes 2D convolution. This is followed by downsampling:}
\label{eq:downsample}
w_i^{(k)}(m,n) &= v_i^{(k)}(s_k m,s_k n),
\intertext{where $s_k$ is the downsampling factor for stage $k$. Each stage concludes with a GDN operation:}
\label{eq:gdn}
u_i^{(k+1)}(m,n) &= \frac {w_i^{(k)}(m,n)} {\Bigl(\beta_{k,i} + \sum_j \gamma_{k,ij} \bigl(w_j^{(k)}(m,n)\bigr)^2 \Bigr)^{\frac 1 2}}.
\end{align}
The full set of $h$, $c$, $\beta$, and $\gamma$ parameters (across all three stages) constitute the parameter vector $\bm \phi$ to be optimized.

Analogously, the synthesis transform $g_s$ consists of three stages, with the order of operations reversed within each stage, downsampling replaced by upsampling, and GDN replaced by an approximate inverse we call IGDN (more details in the appendix). We define $\hat u_i^{(k)}(m,n)$ as the input to the $k$th synthesis stage, such that $\bm{\hat y}$ corresponds to $\hat u_i^{(0)}(m,n)$, and $\bm{\hat x}$ to $\hat u_i^{(3)}(m,n)$. Each stage then consists of the IGDN operation:
\begin{align}
\label{eq:igdn}
\hat w_i^{(k)}(m,n) &= \hat u_i^{(k)}(m,n) \cdot \Bigl(\hat \beta_{k,i} + \sum_j \hat \gamma_{k,ij} \bigl(\hat u_j^{(k)}(m,n)\bigr)^2 \Bigr)^{\frac 1 2},
\intertext{which is followed by upsampling:}
\label{eq:upsample}
\hat v_i^{(k)}(m,n) &= \begin{cases}
\hat w_i^{(k)}(m/\hat s_k,n/\hat s_k) & \text{if } m/\hat s_k \text{ and } n/\hat s_k \text{ are integers,} \\
0 & \text{otherwise,}
\end{cases}
\intertext{where $\hat s_k$ is the upsampling factor for stage $k$. Finally, this is followed by an affine convolution:}
\label{eq:iconv}
\hat u_i^{(k+1)}(m, n) &= \sum_j \bigl(\hat h_{k,ij} \ast \hat v_j^{(k)}\bigr)(m, n) + \hat c_{k,i}.
\end{align}
Analogously, the set of $\hat h$, $\hat c$, $\hat \beta$, and $\hat \gamma$ make up the parameter vector $\bm \theta$. Note that the down\-/upsampling operations can be implemented jointly with their adjacent convolution, improving computational efficiency.

In previous work, we used a perceptual transform $g_p$, separately optimized to mimic human judgements of grayscale image distortions \citep{LaBaBeSi16}, and showed that a set of one-stage transforms optimized for this distortion measure led to visually improved results \citep{BaLaSi16}. Here, we set the perceptual transform $g_p$ to the identity, and use mean squared error (MSE) as the metric (i.e., $d(\bm z, \bm{\hat z})=\|\bm z - \bm{\hat z}\|_2^2$). This allows a more interpretable comparison to existing methods, which are generally optimized for MSE, and also allows optimization for color images, for which we do not currently have a reliable perceptual metric.

%%% ==============================================
\section{Optimization of nonlinear transform coding model}
\label{sec:ntc}

\begin{figure}
%\the\linewidth
\hspace{-1ex}\begin{tikzpicture}[x=1.5cm,y=1.5cm]
\draw[->] (0,0) -- (5,0) node[midway,sloped,below] {$R$};
\draw[->] (0,0) -- (0,3) node[midway,sloped,above] {$D$};
\path[domain=0.7:5,samples=200,fill=black,opacity=.1] plot (\x,{(\x-.2)^(-1)+1}) -- (5,3);
\draw[thick,domain=0.7:5,samples=200] plot (\x,{(\x-.2)^(-1)+1});
\draw[->,domain=3.9:4.9,samples=100] plot (\x,{(\x-.2)^(-1)+1.3}) node[above left,align=right,inner xsep=0,inner ysep=2ex] {generative\\ models \\ $\lambda \to \infty$};
%curve: (\x-.2)^(-1)+1
%slope: -(\x-.2)^(-2)
%inverse: (\y-1)^(-1)+.2
%inverse slope: -(\y-1)^(-2)
\draw[densely dotted] ({(2-1)^(-1)+.2-(2-1)^(-2)*(3-2)},3) -- ({(2-1)^(-1)+.2-(2-1)^(-2)*(0-2)},0) node[sloped,below left,pos=.95] {\footnotesize $R + \lambda_1 D = \const$};
\draw[densely dotted] (0,{(2.5-.2)^(-1)+1 - ((2.5-.2)^(-2)) * (0 - 2.5)}) -- (5,{(2.5-.2)^(-1)+1 - ((2.5-.2)^(-2)) * (5 - 2.5)}) node[sloped,below left,pos=.95] {\footnotesize $R + \lambda_2 D = \const$};
\node[circle,fill=black,minimum size=4pt, inner sep=0pt] at ({(2-1)^(-1)+.2},2) {};
\node[above right,align=left,inner xsep=0,inner ysep=1ex] at ({(2-1)^(-1)+.2},2) {compression\\ model\\ $\lambda = \lambda_1$};
\node[circle,fill=black,minimum size=4pt, inner sep=0pt] at (2.5,{(2.5-.2)^(-1)+1}) {};
\node[above right,align=left,inner xsep=0,inner ysep=1ex] at (2.5,{(2.5-.2)^(-1)+1}) {compression\\ model\\ $\lambda = \lambda_2$};
\end{tikzpicture}\hfill%
\includegraphics[width=.4\linewidth]{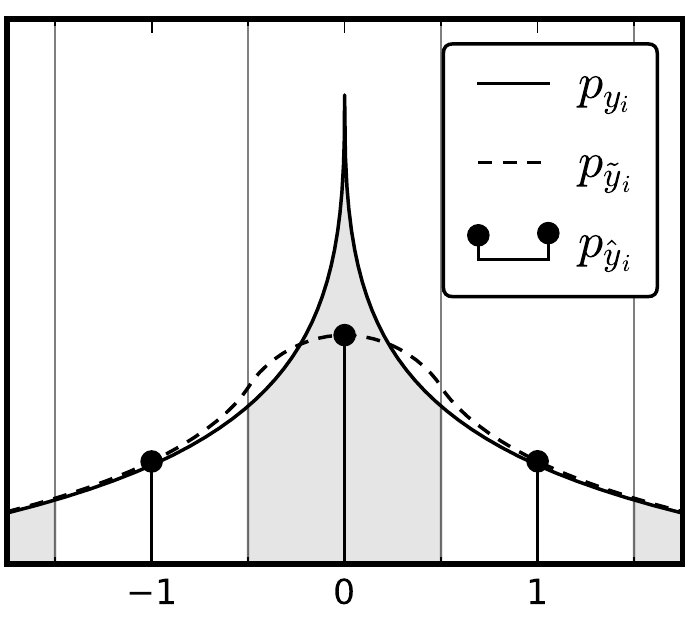}
\caption{Left: The rate--distortion trade-off. The gray region represents the set of all rate--distortion values that can be achieved (over all possible parameter settings). Optimal performance for a given choice of $\lambda$ corresponds to a point on the convex hull of this set with slope $-1/\lambda$. Right: One-dimensional illustration of relationship between densities of $y_i$ (elements of code space), $\hat y_i$ (quantized elements), and $\tilde y_i$ (elements perturbed by uniform noise). Each discrete probability in $p_{\hat y_i}$ equals the probability mass of the density $p_{y_i}$ within the corresponding quantization bin (indicated by shading). The density $p_{\tilde y_i}$ provides a continuous function that interpolates the discrete probability values $p_{\hat y_i}$ at integer positions.}
\label{fig:illustrations}
\end{figure}

Our objective is to minimize a weighted sum of the rate and distortion, $R + \lambda D$, over the parameters of the analysis and synthesis transforms and the entropy code, where $\lambda$ governs the trade-off between the two terms (figure~\ref{fig:illustrations}, left panel). Rather than attempting optimal quantization directly in the image space, which is intractable due to the high dimensionality, we instead assume a fixed uniform scalar quantizer in the code space, and aim to have the nonlinear transformations warp the space in an appropriate way, effectively implementing a parametric form of vector quantization (figure~\ref{fig:overview}). The actual rates achieved by a properly designed entropy code are only slightly larger than the entropy~\citep{RiLa81}, and thus we define the objective functional directly in terms of entropy:
\begin{equation}
\label{eq:L}
L[g_a,g_s,P_{\bm q}] = -\E \bigl[\log_2 P_{\bm q}\bigr] + \lambda \E\bigl[d(\bm z, \bm{\hat z})\bigr],
\end{equation}
where both expectations will be approximated by averages over a training set of images. Given a powerful enough set of transformations, we can assume without loss of generality that the quantization bin size is always one and the representing values are at the centers of the bins. That is,
\begin{equation}
\label{eq:quantizer}
\hat y_i = q_i = \round(y_i),
\end{equation}
where index $i$ runs over all elements of the vectors, including channels and spatial locations. The marginal density of $\hat y_i$ is then given by a train of discrete probability masses (Dirac delta functions, figure~\ref{fig:illustrations}, right panel) with weights equal to the probability mass function of $q_i$:
\begin{equation}
P_{q_i}(n) = \int_{n-\frac 1 2}^{n+\frac 1 2} p_{y_i}(t) \D t, \quad \text{ for all } n \in \Z.
\end{equation}

Note that both terms in \eqref{eq:L} depend on the quantized values, and the derivatives of the quantization function \eqref{eq:quantizer} are zero almost everywhere, rendering gradient descent ineffective. To allow optimization via stochastic gradient descent, we replace the quantizer with an additive i.i.d. uniform noise source $\Delta \bm y$, which has the same width as the quantization bins (one). This relaxed formulation has two desirable properties.
First, the density function of $\bm{\tilde y} = \bm y + \Delta \bm y$ is a continuous relaxation of the probability mass function of $\bm q$ (figure~\ref{fig:illustrations}, right panel):
\begin{equation}
\label{eq:density_relaxation}
p_{\bm{\tilde y}}(\bm n) = P_{\bm q}(\bm n), \quad \text{ for all } \bm n \in \Z^M,
\end{equation}
which implies that the differential entropy of $\bm{\tilde y}$ can be used as an approximation of the entropy of $\bm q$. Second, independent uniform noise approximates quantization error in terms of its marginal moments, and is frequently used as a model of quantization error~\citep{GrNe98}. We can thus use the same approximation for our measure of distortion. We examine the empirical quality of these rate and distortion approximations in section~\ref{sec:experiments}.

We assume independent marginals in the code space for both the relaxed probability model of~$\bm{\tilde y}$ and the entropy code, and model the marginals $p_{\tilde y_i}$ non-parametrically to reduce model error. Specifically, we use finely sampled piecewise linear functions which we update similarly to one-dimensional histograms (see appendix). Since $p_{\tilde y_i} = p_{y_i} \ast \mathcal U(0,1)$ is effectively smoothed by a box-car filter -- the uniform density on the unit interval, $\mathcal U(0,1)$ -- the model error can be made arbitrarily small by decreasing the sampling interval.

Given this continuous approximation of the quantized coefficient distribution, the loss function for parameters $\bm \theta$ and $\bm \phi$ can be written as:
\begin{multline}
\qquad L(\bm \theta, \bm \phi) = \E_{\bm x,\Delta \bm y} \Biggl[ -\sum_i \log_2 p_{\tilde y_i}(g_a(\bm x; \bm \phi) + \Delta \bm y; \bm \psi^{(i)}) \\
+ \lambda\, d\Bigl( g_p\bigl(g_s(g_a(\bm x; \bm \phi) + \Delta \bm y; \bm \theta)\bigr), g_p(\bm x) \Bigr) \Biggr]. \qquad
\label{eq:loss_function}
\end{multline}
where vector $\bm \psi^{(i)}$ parameterizes the piecewise linear approximation of $p_{\tilde y_i}$ (trained jointly with $\bm \theta$ and $\bm \varphi$). This is continuous and differentiable, and thus well-suited for stochastic optimization.

\subsection{Relationship to variational generative image models}
\label{sec:generative_models}

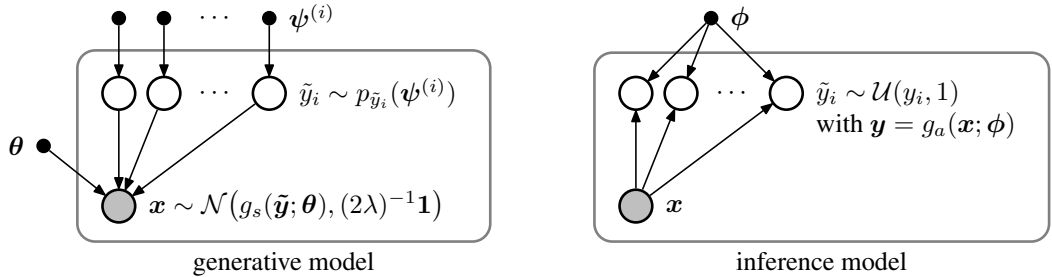
\begin{figure}
\begin{tikzpicture}[x=2cm,y=2cm]
\node [ob] (x) {};
\node [rv] (y0) at ($(x)+(0,.75)$) {};
\node [rv] (y1) at ($(y0)+(.3,0)$) {};
\node [rv] (yn) at ($(y0)+(1,0)$) {};
\node [pa] (theta) at ($(x)+(-.5,.4)$) {};
\node [pa] (psi_0) at ($(y0)+(0,.5)$) {};
\node [pa] (psi_1) at ($(y1)+(0,.5)$) {};
\node [pa] (psi_n) at ($(yn)+(0,.5)$) {};
\node [right=1pt of x] (xlbl) {$\bm x \sim \mathcal N\bigl( g_s( \bm{\tilde y}; \bm \theta ), (2\lambda)^{-1} \bm 1 \bigr)$};
\node [right=1pt of yn] (ylbl) {$\tilde y_i \sim p_{\tilde y_i}(\bm \psi^{(i)})$};
\node [left=1pt of theta] {$\bm \theta$};
\node [right=1pt of psi_n] {$\bm \psi^{(i)}$};
\path (y1) -- node[midway] {$\cdots$} (yn);
\path (psi_1) -- node[midway] {$\cdots$} (psi_n);
\draw[->] (y0) -- (x);
\draw[->] (y1) -- (x);
\draw[->] (yn) -- (x);
\draw[->] (theta) -- (x);
\draw[->] (psi_0) -- (y0);
\draw[->] (psi_1) -- (y1);
\draw[->] (psi_n) -- (yn);
\begin{pgfonlayer}{background}
\node[rounded corners=1.5ex,draw,thick,opacity=.5,fit=(x)(y0)(ylbl),inner xsep=2ex,inner ysep=1.5ex] (image) {};
\end{pgfonlayer}
\node [below=0pt of image,text height=1.5ex, text depth=.25ex] {generative model};
\end{tikzpicture}\hfill%
\begin{tikzpicture}[x=2cm,y=2cm]
\node [ob] (x) {};
\node [rv] (y0) at ($(x)+(0,.75)$) {};
\node [rv] (y1) at ($(y0)+(.3,0)$) {};
\node [rv] (yn) at ($(y0)+(1,0)$) {};
\node [pa] (phi) at ($(y0)+(.5,.5)$) {};
\node [right=1pt of x] (xlbl) {$\bm x$};
\node [right=1pt of yn] (ylbl) {$\tilde y_i \sim \mathcal U( y_i, 1 )$};
\node [right] (ylbl2) at ($(ylbl.west)+(0,-1.2em)$) {with $\bm y = g_a( \bm x; \bm \phi )$};
\node [right=1pt of phi] {$\bm \phi$};
\path (y1) -- node[midway] {$\cdots$} (yn);
\draw[<-] (y0) -- (x);
\draw[<-] (y1) -- (x);
\draw[<-] (yn) -- (x);
\draw[->] (phi) -- (y0);
\draw[->] (phi) -- (y1);
\draw[->] (phi) -- (yn);
\begin{pgfonlayer}{background}
\node[rounded corners=1.5ex,draw,thick,opacity=.5,fit=(x)(y0)(xlbl)(ylbl)(ylbl2),inner xsep=2ex,inner ysep=1.5ex] (image) {};
\end{pgfonlayer}
\node [below=0pt of image,text height=1.5ex, text depth=.25ex] {inference model};
\end{tikzpicture}
\caption{Representation of the relaxed rate--distortion optimization problem as the encoder and decoder graphs of a variational autoencoder. Nodes represent random variables, and gray shading indicates observed data; small filled nodes represent parameters; arrows indicate dependency; and nodes within boxes are per-image.}
\label{fig:autoencoders}
\end{figure}

We derived our formulation directly from the classical rate--distortion optimization problem. However, once the transition to a continuous loss function is made, the optimization problem resembles those encountered in fitting generative models of images, and can more specifically be cast in the context of variational autoencoders~\citep{KiWe14,ReMoWi14}. In Bayesian variational inference, we are given an ensemble of observations of a random variable $x$ along with a generative model $p_{x|y}(x|y)$. We seek to find a posterior $p_{y|x}(y|x)$, which generally cannot be expressed in closed form. The approach followed by \citet{KiWe14} consists of approximating this posterior with a density $q(y|x)$, by minimizing the Kullback--Leibler divergence between the two:
\begin{align}
D_{\mathrm{KL}}\bigl[ q \| p_{y|x} \bigr] &= \E_{y \sim q} \log q(y|x) - \E_{y \sim q} \log p_{y|x}(y|x) \notag \\
&= \E_{y \sim q} \log q(y|x) - \E_{y \sim q} \log p_{x|y}(x|y) - \E_{y \sim q} \log p_y(y) + \const.
\label{eq:kingma_welling}
\end{align}
This objective function is equivalent to our relaxed rate--distortion optimization problem, with distortion measured as MSE, if we define the generative model as follows (figure~\ref{fig:autoencoders}):
\begin{align}
p_{\bm x|\bm{\tilde y}}(\bm x|\bm{\tilde y};\lambda, \bm \theta) &= \mathcal N\bigl( \bm x; g_s(\bm{\tilde y}; \bm \theta), (2\lambda)^{-1} \bm 1\bigr), \label{eq:slack_term} \\
p_{\bm{\tilde y}}(\bm{\tilde y};\bm \psi^{(0)},\bm \psi^{(1)},\dotsc) &= \prod_i p_{\tilde y_i}( \tilde y_i; \bm \psi^{(i)} ),
\intertext{and the approximate posterior as follows:}
q(\bm{\tilde y}|\bm x; \bm \phi) &= \prod_i \mathcal U(\tilde y_i; y_i,1) \quad \text{with } \bm y = g_a( \bm x; \bm \phi ),
\end{align}
where $\mathcal U(\tilde y_i; y_i, 1)$ is the uniform density on the unit interval centered on $y_i$.
With this, the first term in the Kullback--Leibler divergence is constant; the second term corresponds to the distortion, and the third term corresponds to the rate (both up to additive constants). Note that if a perceptual transform $g_p$ is used, or the metric $d$ is not Euclidean, $p_{\bm x|\bm{\tilde y}}$ is no longer Gaussian, and equivalence to variational autoencoders cannot be guaranteed, since the distortion term may not correspond to a normalizable density. For any affine and invertible perceptual transform and any translation-invariant metric, it can be shown to correspond to the density
\begin{equation}
p_{\bm x|\bm{\tilde y}}( \bm x|\bm{\tilde y}; \lambda, \bm \theta ) = \frac 1 {Z(\lambda)} \exp\Biggl( -\lambda\, d\Bigl(g_p\bigl(g_s(\bm{\tilde y}; \bm \theta)\bigr), g_p(\bm x)\Bigr) \Biggr),
\label{eq:distortion_density}
\end{equation}
where $Z(\lambda)$ normalizes the density (but need not be computed to fit the model).

Despite the similarity between our nonlinear transform coding framework and that of variational autoencoders, it is worth noting several fundamental differences. First, variational autoencoders are continuous-valued, and digital compression operates in the discrete domain. Comparing differential entropy with (discrete) entropy, or entropy with an actual bit rate, can potentially lead to misleading results. In this paper, we use the continous domain strictly for optimization, and perform the evaluation on actual bit rates, which allows comparison to existing image coding methods. We assess the quality of the rate and distortion approximations empirically.

Second, generative models aim to minimize differential entropy of the data ensemble under the model, i.e., explaining fluctuations in the data. This often means minimizing the variance of a “slack” term like \eqref{eq:slack_term}, which in turn \emph{maximizes} $\lambda$. Transform coding methods, on the other hand, are optimized to achieve the best trade-off between having the model explain the data (which increases rate and decreases distortion), and having the slack term explain the data (which decreases rate and increases distortion). The overall performance of a compression model is determined by the shape of the convex hull of attainable model distortions and rates, over all possible values of the model parameters. Finding this convex hull is equivalent to optimizing the model for \emph{particular} values of~$\lambda$ (see figure~\ref{fig:illustrations}). In contrast, generative models operate in a regime where $\lambda$ is inferred and ideally approaches infinity for noiseless data, which corresponds to the regime of lossless compression. Even so, lossless compression methods still need to operate in a discretized space, typically directly on quantized luminance values. For generative models, the discretization of luminance values is usually considered a nuisance~\citep{ThOoBe15}, although there are examples of generative models that operate on quantized pixel values \citep{OoKaKa16}.

Finally, although correspondence between the typical slack term \eqref{eq:slack_term} of a generative model (figure~\ref{fig:autoencoders}, left panel) and the distortion metric in rate--distortion optimization holds for simple metrics (e.g., Euclidean distance), a more general perceptual measure would be considered a peculiar choice from a generative modeling perspective, if it corresponds to a density at all.

%%% ==============================================
\section{Experimental results}
\label{sec:experiments}%
We jointly optimized the full set of parameters $\bm \phi$, $\bm \theta$, and all $\bm \psi$ over a subset of the ImageNet database~\citep{DeDoSoLiLi09} consisting of 6507 images using stochastic descent. This optimization was performed separately for each $\lambda$, yielding separate transforms and marginal probability models for each value.

For the grayscale analysis transform, we used 128 filters (size $9\times 9$) in the first stage, each subsampled by a factor of 4 vertically and horizontally. The remaining two stages retain the number of channels, but use filters operating across all input channels ($5 \times 5 \times 128$), with outputs subsampled by a factor of 2 in each dimension. The net output thus has half the dimensionality of the input. The synthesis transform is structured analogously. For RGB images, we trained a separate set of models, with the first stage augmented to operate across three (color) input channels. For the two largest values of $\lambda$, and for RGB models, we increased the network capacity by increasing the number of channels in each stage to 256 and 192, respectively. Further details about the parameterization of the transforms and their training can be found in the appendix.

\begin{figure}
\centering%
\includegraphics[width=.45\linewidth,trim=8 25 10 10]{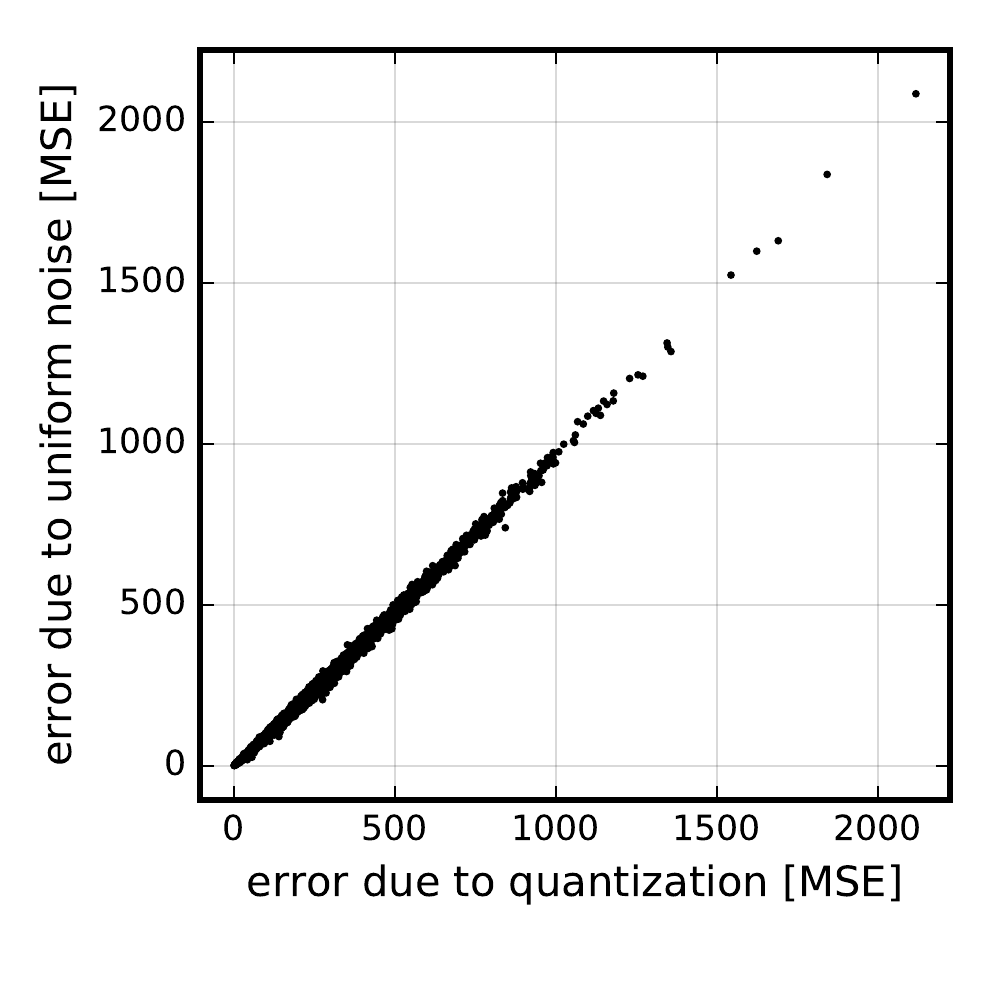}\hfill%
\includegraphics[width=.45\linewidth,trim=8 25 10 10]{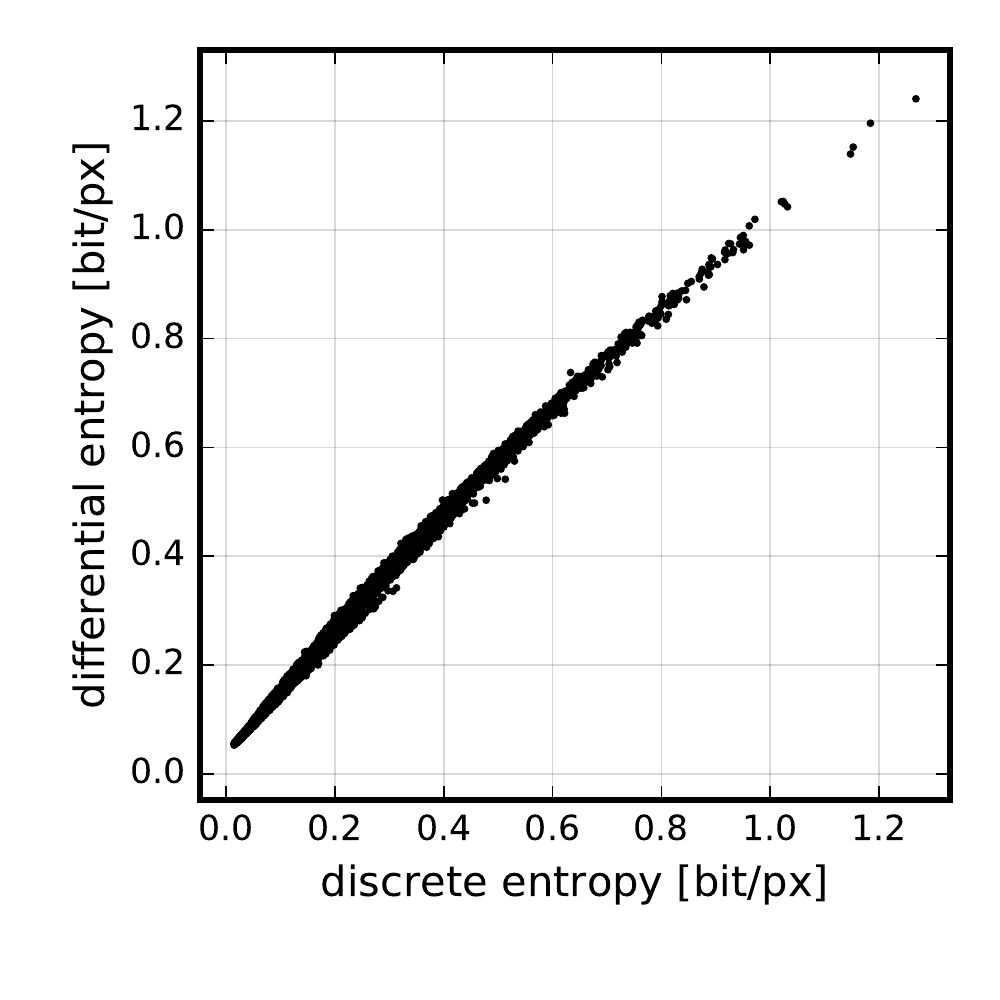}%
\caption{Scatter plots comparing discrete vs. continuously-relaxed values of the two terms of the objective function, evaluated for the optimized GDN model.
Points correspond to different values of $\lambda$ between 32 and 2048 (inclusive), for images drawn from a random subset of 2169 images (one third) from the training set. Left: distortion term, evaluated for $g_s(\bm{\hat y})$ vs. $g_s(\bm{\tilde y})$. Right: rate term, $H[P_{q_i}]$ vs. $h[p_{\tilde y_i}]$ (summed over $i$).}
\label{fig:objective}
\end{figure}

We first verified that the continuously-relaxed loss function given in section~\ref{sec:ntc} provides a good approximation to the actual rate--distortion values obtained with quantization (figure~\ref{fig:objective}). The relaxed distortion term appears to be mostly unbiased, and exhibits a relatively small variance. The relaxed (differential) entropy provides a somewhat positively biased estimate of the discrete entropy for the coarser quantization regime, but the bias disappears for finer quantization, as expected. Note that since the values of $\lambda$ do not have any intrinsic meaning, but serve only to map out the convex hull of optimal points in the rate--distortion plane (figure~\ref{fig:illustrations}, left panel), a constant bias in either of the terms would simply alter the effective value of $\lambda$, with no effect on the compression performance.

\begin{figure}
\centering\footnotesize%
\vspace{-1.5em}\includegraphics[width=\textwidth,height=.3\textheight,keepaspectratio]{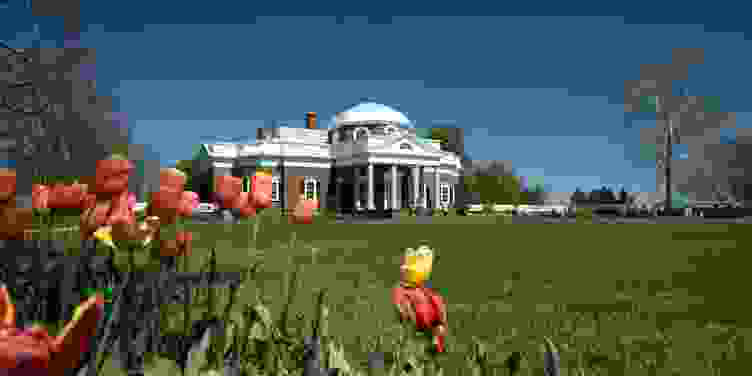}\\%
{\bf JPEG}, 4283 bytes (0.121 bit/px), PSNR: luma 24.85 dB/chroma 29.23 dB, MS-SSIM: 0.8079\vspace{3pt}\\%
\includegraphics[width=\textwidth,height=.3\textheight,keepaspectratio]{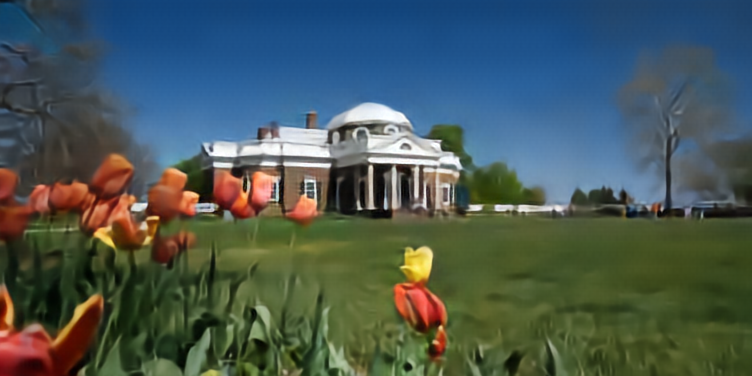}\\%
{\bf Proposed method}, 3986 bytes (0.113 bit/px), PSNR: luma 27.01 dB/chroma 34.16 dB, MS-SSIM: 0.9039\vspace{3pt}\\%
\includegraphics[width=\textwidth,height=.3\textheight,keepaspectratio]{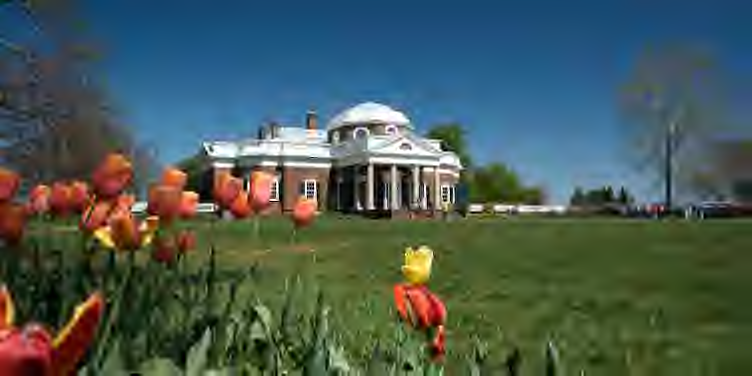}\\%
{\bf JPEG 2000}, 4004 bytes (0.113 bit/px), PSNR: luma 26.61 dB/chroma 33.88 dB, MS-SSIM: 0.8860%
\caption{A heavily compressed example image, $752\times 376$ pixels. Note the appearance of artifacts, especially near edges, in both the JPEG and JPEG2000 images.\vspace{-1.5em}}%
\label{fig:example}
\end{figure}

\begin{figure}
\centering\footnotesize%
\includegraphics[width=.33\textwidth,trim=0 16 0 0,clip]{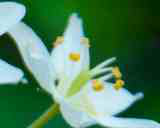}\hfill%
\includegraphics[width=.33\textwidth,trim=0 16 0 0,clip]{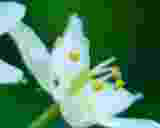}\hfill%
\includegraphics[width=.33\textwidth,trim=0 16 0 0,clip]{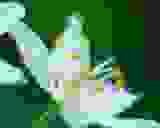}\vspace{1.5pt}\\%
\includegraphics[width=.33\textwidth,trim=0 16 0 0,clip]{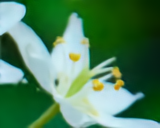}\hfill%
\includegraphics[width=.33\textwidth,trim=0 16 0 0,clip]{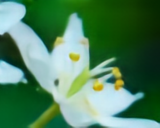}\hfill%
\includegraphics[width=.33\textwidth,trim=0 16 0 0,clip]{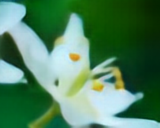}\vspace{1.5pt}\\%
\includegraphics[width=.33\textwidth,trim=0 16 0 0,clip]{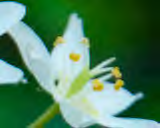}\hfill%
\includegraphics[width=.33\textwidth,trim=0 16 0 0,clip]{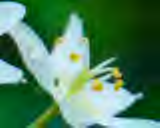}\hfill%
\includegraphics[width=.33\textwidth,trim=0 16 0 0,clip]{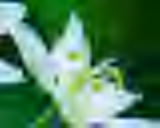}%
\caption{Cropped portion of an image compressed at three different bit rates. Middle row: the proposed method, at three different settings of $\lambda$. Top row: JPEG, with three different quality settings. Bottom row: JPEG~2000, with three different rate settings. Bit rates within each column are matched.}
\label{fig:degradation}
\end{figure}

\begin{figure}
\centering%
\includegraphics[width=.5\linewidth]{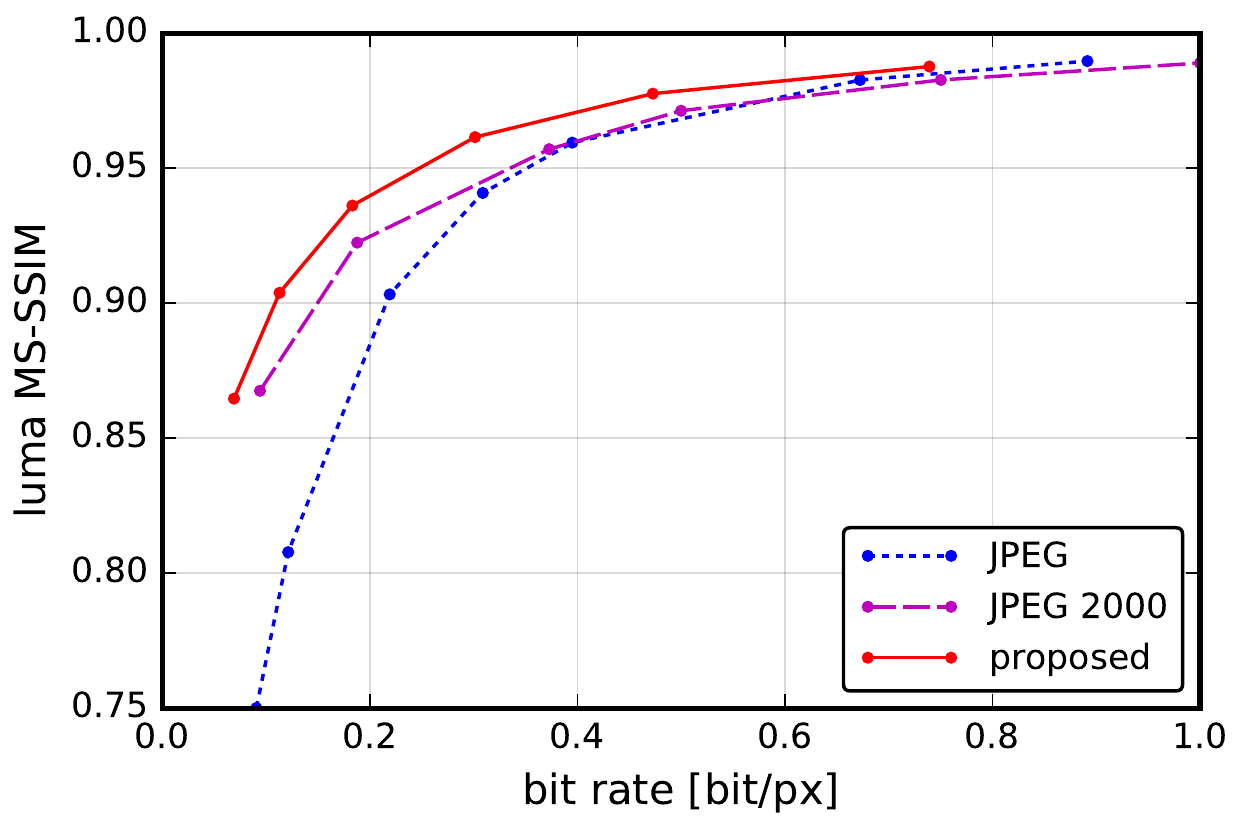}\hfill%
\includegraphics[width=.5\linewidth]{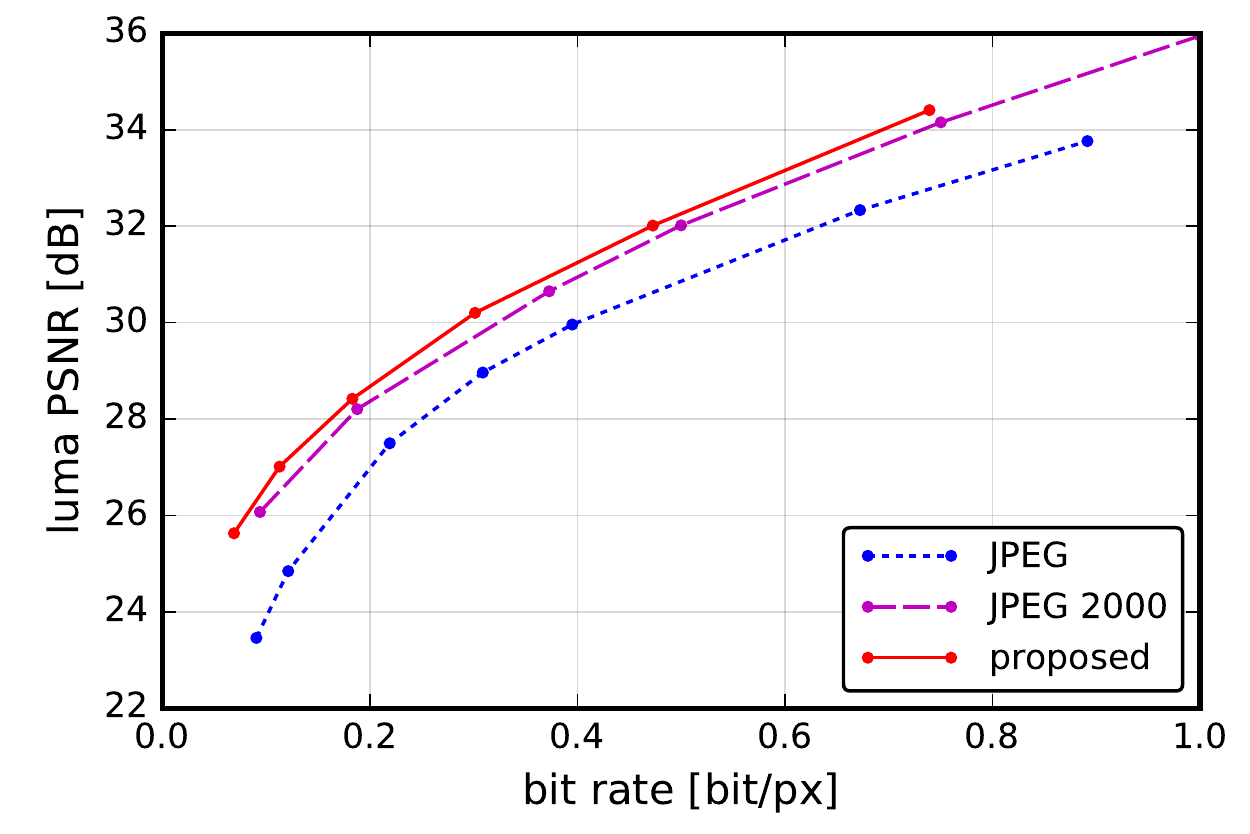}%
\caption{Rate--distortion curves for the luma component of image shown in figure~\ref{fig:example}. Left: perceptual quality, measured with multi-scale structural similarity (MS-SSIM; \citet{WaSiBo03}). Right: peak signal-to-noise ratio ($10 \log_{10} (255^2 /\mathrm{MSE})$).}%
\label{fig:rd-plot}
\end{figure}

We compare the rate--distortion performance of our method to two standard methods: JPEG and JPEG~2000. For our method, all images were compressed using uniform quantization (the continuous relaxation using additive noise was used only for training purposes). To make the comparisons more fair, we implemented a simple entropy code based on the context-based adaptive binary arithmetic coding framework (CABAC; \citealp{MaScWi03}). All sideband information needed by the decoder (size of images, value of $\lambda$, etc.) was included in the bit stream (see appendix).
Note that although the computational costs for training our models are quite high, encoding or decoding an image with the trained models is efficient, requiring only execution of the optimized analysis transformation and quantizer, or the synthesis transformation, respectively.
Evaluations were performed on the Kodak image dataset\footnote{Downloaded from \url{http://www.cipr.rpi.edu/resource/stills/kodak.html}}, an uncompressed set of images commonly used to evaluate image compression methods.
We also examined a set of relatively standard (if outdated) images used by the compression community (known by the names “Lena”, “Barbara”, “Peppers”, and “Mandrill”) as well as a set of our own digital photographs. None of these test images was included in the training set. All test images, compressed at a variety of bit rates using all three methods, along with their associated rate--distortion curves, are available online at \url{http://www.cns.nyu.edu/\~lcv/iclr2017}.

Although we used MSE as a distortion metric for training, the appearance of compressed images is both qualitatively different and substantially improved, compared to JPEG and JPEG~2000. As an example, figure~\ref{fig:example} shows an image compressed using our method optimized for a low value of $\lambda$ (and thus, a low bit rate), compared to JPEG/JPEG~2000 images compressed at equal or greater bit rates. The image compressed with our method has less detail than the original (not shown, but available online), with fine texture and other patterns often eliminated altogether, but this is accomplished in a way that preserves the smoothness of contours and sharpness of many of the edges, giving them a natural appearance. By comparison, the JPEG and JPEG~2000 images exhibit artifacts that are common to all linear transform coding methods: since local features (edges, contours, texture elements, etc.) are represented using particular combinations of localized linear basis functions, independent scalar quantization of the transform coefficients causes imbalances in these combinations, and leads to visually disturbing blocking, aliasing, and ringing artifacts that reflect the underlying basis functions.

Remarkably, we find that the perceptual advantages of our method hold for \emph{all} images tested, and at all bit rates. The progression from high to low bit rates is shown for an example image in figure~\ref{fig:degradation} (additional examples provided in appendix and online). As bit rate is reduced, JPEG and JPEG~2000 degrade their approximation of the original image by coarsening the precision of the coefficients of linear basis functions, thus exposing the visual appearance of those basis functions. On the other hand, our method appears to progressively simplify contours and other image features, effectively concealing the underlying quantization of the representation. Consistent with the appearance of these example images, we find that distortion measured with a perceptual metric (MS-SSIM; \citealp{WaSiBo03}), indicates substantial improvements across all tested images and bit rates (figure \ref{fig:rd-plot}; additional examples provided in the appendix and online). Finally, when quantified with PSNR, we find that our method exhibits better rate--distortion performance than both JPEG and JPEG~2000 for most (but not all) test images, especially at the lower bit rates.

%%% ==============================================
\section{Discussion}
We have presented a complete image compression method based on \emph{nonlinear transform coding}, and a framework to optimize it end-to-end for rate--distortion performance. Our compression method offers improvements in rate--distortion performance over JPEG and JPEG~2000 for most images and bit rates. More remarkably, although the method was optimized using mean squared error as a distortion metric, the compressed images are much more natural in appearance than those compressed with JPEG or JPEG~2000, both of which suffer from the severe artifacts commonly seen in linear transform coding methods. Consistent with this, perceptual quality (as estimated with the MS-SSIM index) exhibits substantial improvement across all test images and bit rates. We believe this visual improvement arises because the cascade of biologically-inspired nonlinear transformations in the model have been optimized to capture the features and attributes of images that are represented in the statistics of the data, parallel to the processes of evolution and development that are believed to have shaped visual representations within the human brain \citep{SiOl01}. Nevertheless, additional visual improvements might be possible if the method were optimized using a perceptual metric in place of MSE \citep{BaLaSi16}.

For comparison to linear transform coding methods, we can interpret our analysis transform as a single-stage linear transform followed by a complex vector quantizer. As in many other optimized representations -- e.g., sparse coding \citep{LeOl98} -- as well as many engineered representations -- e.g., the steerable pyramid \citep{SiFrAdHe92}, curvelets \citep{CaDo02}, and dual-tree complex wavelets \citep{SeBaKi05} -- the filters in this first stage are localized and oriented and the representation is \emph{overcomplete}. Whereas most transform coding methods use complete (often orthogonal) linear transforms with spatially separable filters, the overcompleteness and orientation tuning of our initial transform may explain the ability of the model to better represent features and contours with continuously varying orientation, position and scale \citep{SiFrAdHe92}.

Our work is related to two previous publications that optimize image representations with the goal of image compression.
\citet{GrBeJiDaWi16} introduce an interesting hierarchical representation of images, in which degradations are more natural looking than those of linear representations. However, rather than optimizing directly for rate--distortion performance, their modeling is generative. Due to the differences between these approaches (as outlined in section~\ref{sec:generative_models}), their procedure of obtaining coding representations from the generative model (scalar quantization, and elimination of hierarchical levels of the representation) is less systematic than our approach and unlikely to be optimal. Further, no entropy code is provided, and the authors therefore resort to comparing entropy estimates to bit rates of established compression methods, which can be unreliable. The model developed by \citet{ToViJoHwMi16} is optimized to provide various rate--distortion trade-offs and directly output a binary representation, making it more easily comparable to other image compression methods. Moreover, their formulation has the advantage over ours that a single representation is sought for all rate points. However, it is not clear whether their formulation necessarily leads to rate--distortion optimality (and their empirical results suggest that this is not the case).

We are currently testing models that use simpler rectified-linear or sigmoidal nonlinearities, to determine how much of the performance and visual quality of our results is due to use of biologically-inspired joint nonlinearities. Preliminary results indicate that qualitatively similar results are achievable with other activation functions we tested, but that rectified linear units generally require a substantially larger number of model parameters/stages to achieve the same rate--distortion performance as the GDN/IGDN nonlinearities. This suggests that GDN/IGDN transforms are more efficient for compression, producing better models with fewer stages of processing (as we previously found for density estimation; \citealp{BaLaSi15}), which might be an advantage for deployment of our method, say, in embedded systems. However, such conclusions are based on a somewhat limited set of experiments and should at this point be considered provisional. More generally, GDN represents a multivariate generalization of a particular type of sigmoidal function. As such, the observed efficiency advantage relative to pointwise nonlinearities is expected, and a variant of a universal function approximation theorem (e.g., \citealp{LeLiPiSc93}) should hold.

The rate--distortion objective can be seen as a particular instantiation of the general unsupervised learning or density estimation problems. Since the transformation to a discrete representation may be viewed as a form of classification, it is worth considering whether our framework offers any insights that might be transferred to more specific supervised learning problems, such as object recognition. For example, the additive noise used in the objective function as a relaxation of quantization might also serve the purpose of making supervised classification networks more robust to small perturbations, and thus allow them to avoid catastrophic ``adversarial'' failures that have been demonstrated in previous work \citep{SzZaSuBrEr13}. In any case, our results provide a strong example of the power of end-to-end optimization in achieving a new solution to a classical problem.

\subsubsection*{Acknowledgments}
We thank Olivier Hénaff and Matthias Bethge for fruitful discussions.

\printbibliography
%\bibliography{main}
%\bibliographystyle{iclr2017_conference}

%%% ==============================================
\newpage
\section{Appendix}

\subsection{Network architecture and optimization}
\label{sec:architecture}

\begin{figure}[b]
\begin{tikzpicture}
\node [matrix, inner sep=0pt, column sep=2pt, row sep=1ex,
layer/.style={rectangle, draw, fill=white, rotate=90, minimum width=25ex, inner ysep=5pt},
numpars/.style={rotate=90, font=\footnotesize, left}] (analysis) {
\node[layer] {conv $|\; 9\times 9 \;|\; 128\times 1$}; &
\node[layer] {downsample 4}; &
\node[layer,fill=black!10] {GDN}; &
\node[layer] {conv $|\; 5\times 5 \;|\; 128\times 128$}; &
\node[layer] {downsample 2}; &
\node[layer,fill=black!10] {GDN}; &
\node[layer] {conv $|\; 5\times 5 \;|\; 128\times 128$}; &
\node[layer] {downsample 2}; &
\node[layer,fill=black!10] {GDN}; \\
\node[numpars] {10368}; &
\node[numpars] {0}; &
\node[numpars] {8384}; &
\node[numpars] {409600}; &
\node[numpars] {0}; &
\node[numpars] {8384}; &
\node[numpars] {409600}; &
\node[numpars] {0}; &
\node[numpars] {8384}; \\
};
\node at ($(analysis.north) + (0,4ex)$) (analbl) {analysis};
\draw[->] (analysis.west |- analbl.south) -- (analysis.east |- analbl.south);
\end{tikzpicture}\hfill%
\begin{tikzpicture}
\node [matrix, inner sep=0pt, column sep=2pt, row sep=1ex,
layer/.style={rectangle, draw, fill=white, rotate=90, minimum width=25ex, inner ysep=5pt},
numpars/.style={rotate=90, font=\footnotesize}, left] (synthesis) {
\node[layer,fill=black!10] {IGDN}; &
\node[layer] {upsample 2}; &
\node[layer] {conv $|\; 5\times 5 \;|\; 128\times 128$}; &
\node[layer,fill=black!10] {IGDN}; &
\node[layer] {upsample 2}; &
\node[layer] {conv $|\; 5\times 5 \;|\; 128\times 128$}; &
\node[layer,fill=black!10] {IGDN}; &
\node[layer] {upsample 4}; &
\node[layer] {conv $|\; 9\times 9 \;|\; 1\times 128$}; \\
\node[numpars] {8384}; &
\node[numpars] {0}; &
\node[numpars] {409600}; &
\node[numpars] {8384}; &
\node[numpars] {0}; &
\node[numpars] {409600}; &
\node[numpars] {8384}; &
\node[numpars] {0}; &
\node[numpars] {10241}; \\
};
\node at ($(synthesis.north) + (0,4ex)$) (synlbl) {synthesis};
\draw[->] (synthesis.west |- synlbl.south) -- (synthesis.east |- synlbl.south);
\end{tikzpicture}
\caption{Parameterization of analysis ($g_a$) and synthesis ($g_s$) transforms for grayscale images. \emph{conv}: affine convolution \eqref{eq:conv}/\eqref{eq:iconv}, with filter support ($x\times y$) and number of channels (output$\times$input). \emph{down-/upsample}: regular down-/upsampling \eqref{eq:downsample}/\eqref{eq:upsample} by given factor (implemented jointly with the adjacent convolution). \emph{GDN/IGDN}: generalized divisive normalization across channels \eqref{eq:gdn}, and its approximate inverse \eqref{eq:igdn}; see text. Number of parameters for each layer given at the bottom.}
\label{fig:architecture}
\end{figure}
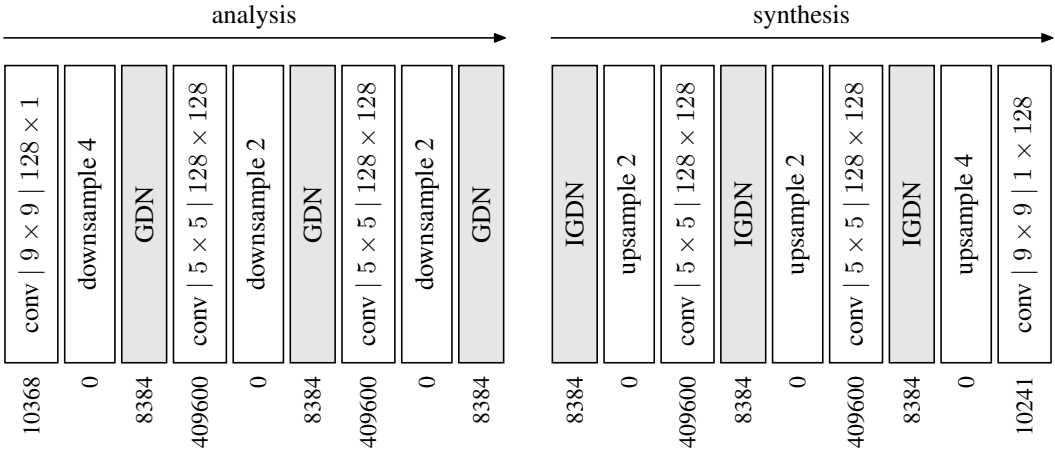

As described in the main text, our analysis transform consists of three stages of convolution, downsampling, and GDN. The number and size of filters, downsampling factors, and connectivity from layer to layer are provided in figure~\ref{fig:architecture} for the grayscale transforms. The transforms for RGB images and for high bit rates differ slightly in that they have an increased number of channels in each stage. These choices are somewhat ad-hoc, and a more thorough exploration of alternative architectures could potentially lead to significant performance improvements.

We have previously shown that GDN is highly efficient in Gaussianizing the local joint statistics of natural images~\citep{BaLaSi15}. Even though Gaussianization is a quite different optimization problem than the rate--distortion objective with the set of constraints defined above, it is similar in that a marginally independent latent model is assumed in both cases. When optimizing for Gaussianization, the exponents in the parametric form of GDN control the tail behavior of the Gaussianized densities. Since tail behavior is less important here, we chose to simplify the functional form, fixing the exponents as well as forcing the weight matrix to be symmetric (i.e., $\gamma_{k,ij}=\gamma_{k,ji}$).

The synthesis transform is meant to function as an approximate inverse transformation, so we construct it by applying a principle known from the LISTA algorithm \citep{GrLe10} to the fixed point iteration previously used to invert the GDN transform~\citep{BaLaSi15}. The approximate inverse consists of one iteration, but with a separate set of parameters from the forward transform, which are constrained in the same way, but trained separately. We refer to this nonlinear transform as “inverse GDN” (IGDN).

The full model (analysis and synthesis filters, GDN and IGDN parameters) were optimized, for each $\lambda$, over a subset of the ImageNet database~\citep{DeDoSoLiLi09} consisting of 6507 images. We applied a number of preprocessing steps to the images in order to reduce artifacts and other unwanted contaminations: first, we eliminated images with excessive saturation. We added a small amount of uniform noise, corresponding to the quantization of pixel values, to the remaining images. Finally, we downsampled and cropped the images to a size of $256\times 256$ pixels each, where the amount of downsampling and cropping was randomized, but depended on the size of the original image. In order to reduce high-frequency noise and compression artifacts, we only allowed resampling factors less than 0.75, discarding images that were too small to satisfy this constraint.

To ensure efficient and stable optimization, we used the following techniques:
\begin{itemize}
\item We used the Adam optimization algorithm~\citep{KiBa14} to obtain values for the parameters $\bm \phi$ and $\bm \theta$, starting with $\alpha = 10^{-4}$, and subsequently lowering it by a factor of~10 whenever the improvement of both rate and distortion stagnated, until $\alpha = 10^{-7}$.
\item Linear filters were parameterized using their discrete cosine transform (DCT) coefficients. We found this to be slightly more effective in speeding up the convergence than discrete Fourier transform (DFT) parameterization \citep{RiSnAd15}.
\item We parameterized the GDN parameters in terms of the elementwise relationship
\begin{equation*}
\beta_{k,i} = (\beta_{k,i}')^2 - 2^{-10}.
\end{equation*}
The squaring ensures that gradients are smaller around parameter values close to 0, a regime in which the optimization can otherwise become unstable. To obtain an unambiguous mapping, we projected each $\beta_{k,i}'$ onto the interval $[2^{-5},\infty)$ after each gradient step. We applied the same treatment to $\gamma_{k,ij}$, and additionally averaged $\gamma'_{k,ij}$ with its transpose after each step in order to make it symmetric as explained above. The IGDN parameters were treated in the same way.
\item To remove the scaling ambiguity between the each linear transform and its following nonlinearity (or preceding nonlinearity, in the case of the synthesis transform), we re-normalized the linear filters after each gradient step, dividing each filter by the square root of the sum of its squared coefficients. For the analysis transform, the sum runs over space and all input channels, and for the synthesis transform, over space and all output channels.
\end{itemize}
We represented each of the marginals $p_{\tilde y_i}$ as a piecewise linear function (i.e., a linear spline), using 10 sampling points per unit interval. The parameter vector $\bm \psi^{(i)}$ consists of the value of $p_{\tilde y_i}$ at these sampling points. We did not use Adam to update $\bm \psi^{(i)}$; rather, we used ordinary stochastic gradient descent to minimize the negative expected likelihood:
\begin{equation}
L_{\psi}( \bm \psi^{(0)}, \bm \psi^{(1)}, \dotsc ) = -\E_{\bm{\tilde y}} \sum_i p_{\tilde y_i}(\tilde y_i; \bm \psi^{(i)}).
\end{equation}
and renormalized the marginal densities after each step. After every $10^6$ gradient steps, we used a heuristic to adapt the range of the spline approximation to cover the range of values of $\tilde y_i$ obtained on the training set.

\subsection{Entropy code}
We implemented an entropy code based on the context-adaptive binary arithmetic coding (CABAC) framework defined by \citet{MaScWi03}. Arithmetic entropy codes are designed to compress discrete-valued data to bit rates closely approaching the entropy of the representation, assuming that the probability model used to design the code describes the data well. The following information was encoded into the bitstream:
\begin{itemize}
\item the size of the image (two 16-bit integers, bypassing arithmetic coding),
\item whether the image is grayscale or RGB (one bit, bypassing arithmetic coding),
\item the value of $\lambda$ (one 16-bit integer, bypassing arithmetic coding), which provides an index for the parameters of the analysis and synthesis transforms as well as the initial probability models for the entropy codes (these are fixed after optimization, and assumed to be available to encoder and decoder).
\item the value of each element of $\bm q$, iterating over channels, and over space in raster-scan order, using the arithmetic coding engine.
\end{itemize}

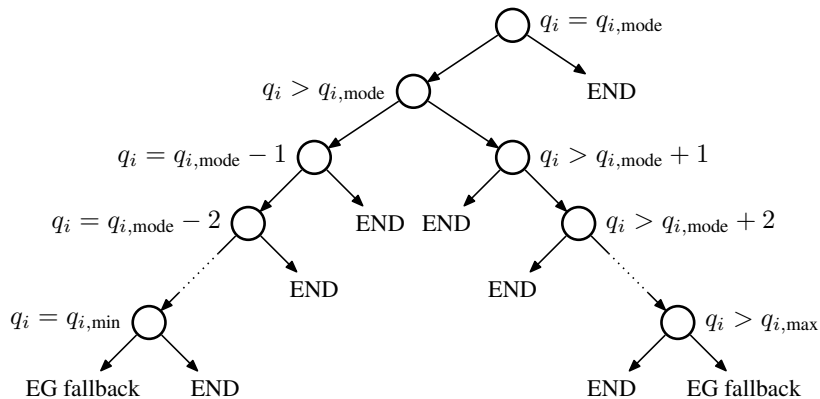
\begin{figure}[t]
\centering%
\begin{tikzpicture}[x=2.5em,y=2.5em]
\node [rv] (zero) {};
\node [right=0 of zero] {$q_i = q_{i,\textrm{mode}}$};

\node [rv] (sign) at ($(zero)+(-1.5,-1)$) {} edge[<-] (zero);
\node [left=0 of sign] {$q_i > q_{i,\textrm{mode}}$};

\node at ($(zero)+(1.5,-1)$) {\footnotesize END} edge[<-] (zero);

\node [rv] (m1) at ($(sign)+(-1.5,-1)$) {} edge[<-] (sign);
\node [left=0 of m1] {$q_i = q_{i,\textrm{mode}} - 1$};

\node [rv] (m2) at ($(m1)+(-1,-1)$) {} edge[<-] (m1);
\node [left=0 of m2] {$q_i = q_{i,\textrm{mode}} - 2$};

\node at ($(m1)+(1,-1)$) {\footnotesize END} edge[<-] (m1);

\node [rv] (m3) at ($(m2)+(-1.5,-1.5)$) {};
\node [left=0 of m3] {$q_i = q_{i,\textrm{min}}$};
\draw (m2) -- ($(m2)!.3!(m3)$);
\draw [dotted] ($(m2)!.3!(m3)$) -- ($(m2)!.7!(m3)$);
\draw [->] ($(m2)!.7!(m3)$) -- (m3);

\node at ($(m2)+(1,-1)$) {\footnotesize END} edge[<-] (m2);

\node at ($(m3)+(-1,-1)$) {\footnotesize EG fallback} edge[<-] (m3);

\node at ($(m3)+(1,-1)$) {\footnotesize END} edge[<-] (m3);

\node [rv] (p1) at ($(sign)+(1.5,-1)$) {} edge[<-] (sign);
\node [right=0 of p1] {$q_i > q_{i,\textrm{mode}} + 1$};

\node at ($(p1)+(-1,-1)$) {\footnotesize END} edge[<-] (p1);

\node [rv] (p2) at ($(p1)+(1,-1)$) {} edge[<-] (p1);
\node [right=0 of p2] {$q_i > q_{i,\textrm{mode}} + 2$};

\node at ($(p2)+(-1,-1)$) {\footnotesize END} edge[<-] (p2);

\node [rv] (p3) at ($(p2)+(1.5,-1.5)$) {};
\node [right=0 of p3] {$q_i > q_{i,\textrm{max}}$};
\draw (p2) -- ($(p2)!.3!(p3)$);
\draw [dotted] ($(p2)!.3!(p3)$) -- ($(p2)!.7!(p3)$);
\draw [->] ($(p2)!.7!(p3)$) -- (p3);

\node at ($(p3)+(-1,-1)$) {\footnotesize END} edge[<-] (p3);

\node at ($(p3)+(1,-1)$) {\footnotesize EG fallback} edge[<-] (p3);
\end{tikzpicture}
\caption{Binarization of a quantized value for binary arithmetic coding. Each circle represents a binary decision encoded with its own CABAC context. Arrows pointing left represent “false”, arrows pointing right “true”. On reaching {\footnotesize END}, the encoding of the quantized value is completed. On reaching {\footnotesize EG fallback}, the magnitude of $q_i$ which falls outside of the range $[q_{i,\textrm{min}},q_{i,\textrm{max}}]$ is encoded using an exponential Golomb code, bypassing the arithmetic coding engine.}
\label{fig:binarization}
\end{figure}

\begin{figure}[t]
\centering%
\includegraphics[width=.5\textwidth]{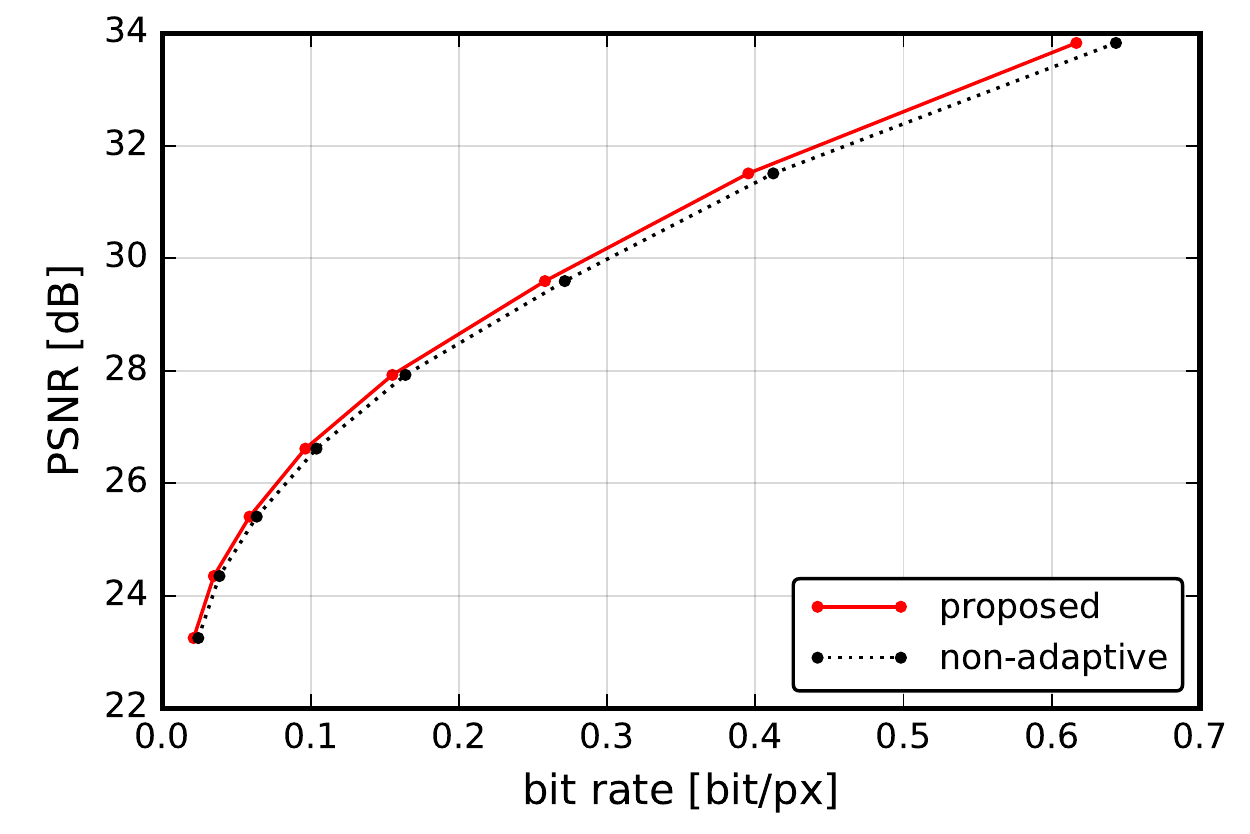}
\caption{Rate--distortion comparison of adaptive vs. non-adaptive entropy coding, averaged (for each value of $\lambda$) over the 24 images in the Kodak test set. The non-adaptive entropy code is simulated by computing the entropy of $\bm q$ assuming the probability model determined during optimization (which is also used to initialize the adaptive code).}
\label{fig:adaptive_coding}
\end{figure}

Since CABAC operates on binary values, the quantized values in $\bm q$ need to be converted to binary decisions. We follow a simple scheme inspired by the encoding of H.264/AVC transform coefficients as detailed by \citet{MaScWi03}. For each $q_i$, we start by testing if the encoded value is equal to the mode of the distribution. If this is the case, the encoding of $q_i$ is completed. If not, another binary decision determines whether it is smaller or larger than the mode. Following that, each possible integer value is tested in turn, which yields a bifurcated chain of decisions as illustrated in figure~\ref{fig:binarization}. This process is carried out until either one of the binary decisions determines $q_i$, or some minimum ($q_{i,\textrm{min}}$) or maximum ($q_{i,\textrm{max}}$) value is reached. In case $q_i$ is outside of that range, the difference between it and the range bound is encoded using an exponential Golomb code, bypassing the arithmetic coding engine.

Adaptive codes, such as CABAC, can potentially further improve bit rates, and to some extent correct model error, by adapting the probability model on-line to the statistics of the data. In our code, this is achieved by sharing the marginal probability model $P_{q_i}$ of each element in $\bm q$ across space within each channel. We derived the initial probability models by subsampling the continuous densities $p_{\tilde y_i}$ determined during optimization, as in \eqref{eq:density_relaxation}. However, note that due to the simple raster-scan ordering, the coding scheme presented above only crudely exploits spatial adaptation of the probability model compared to existing coding methods such as JPEG~2000 and H.264/AVC. Thus, the performance gains compared to a well-designed non-adaptive entropy code are relatively small (figure \ref{fig:adaptive_coding}), and likely smaller than those achieved by the entropy code in JPEG~2000, to which we compare.

\subsection{Evaluation details and additional example images}
\begin{figure}[p]
\centering\footnotesize%
\includegraphics[width=.5\textwidth]{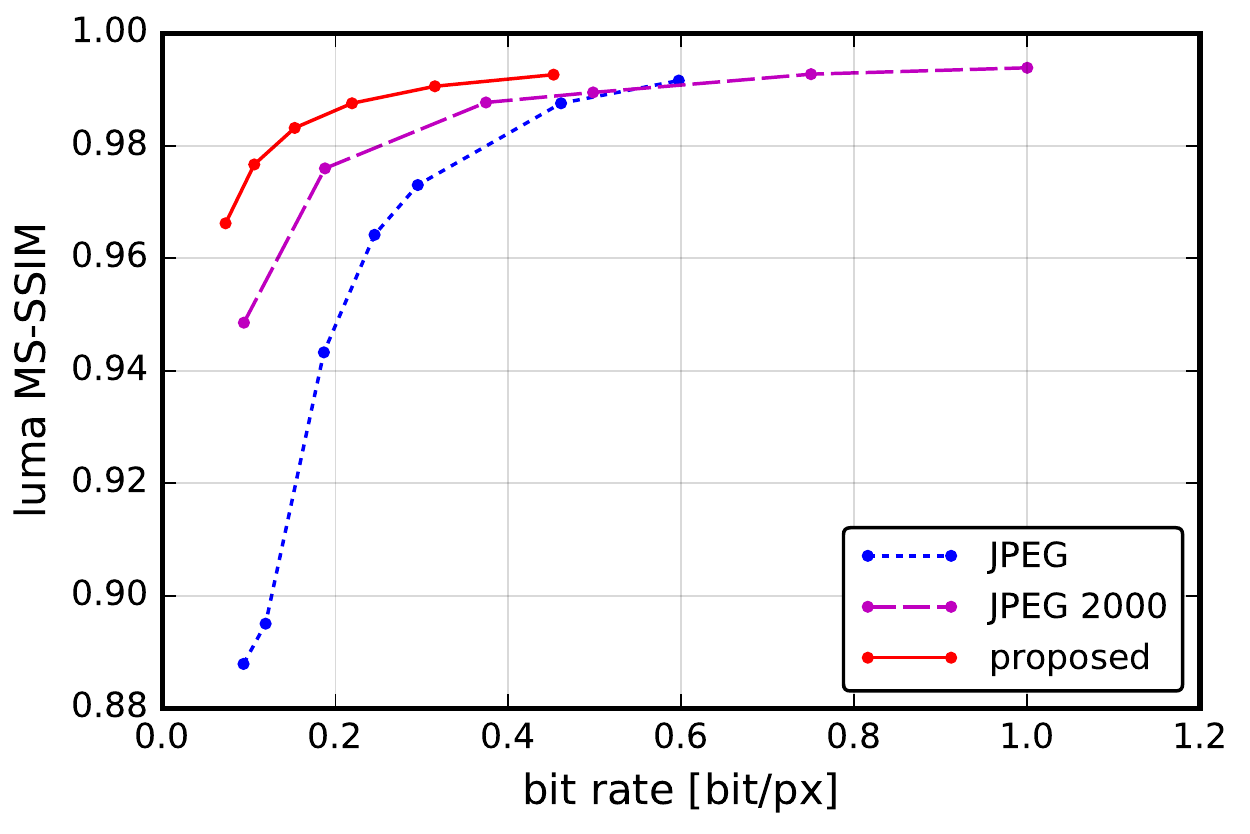}\hfill%
\includegraphics[width=.5\textwidth]{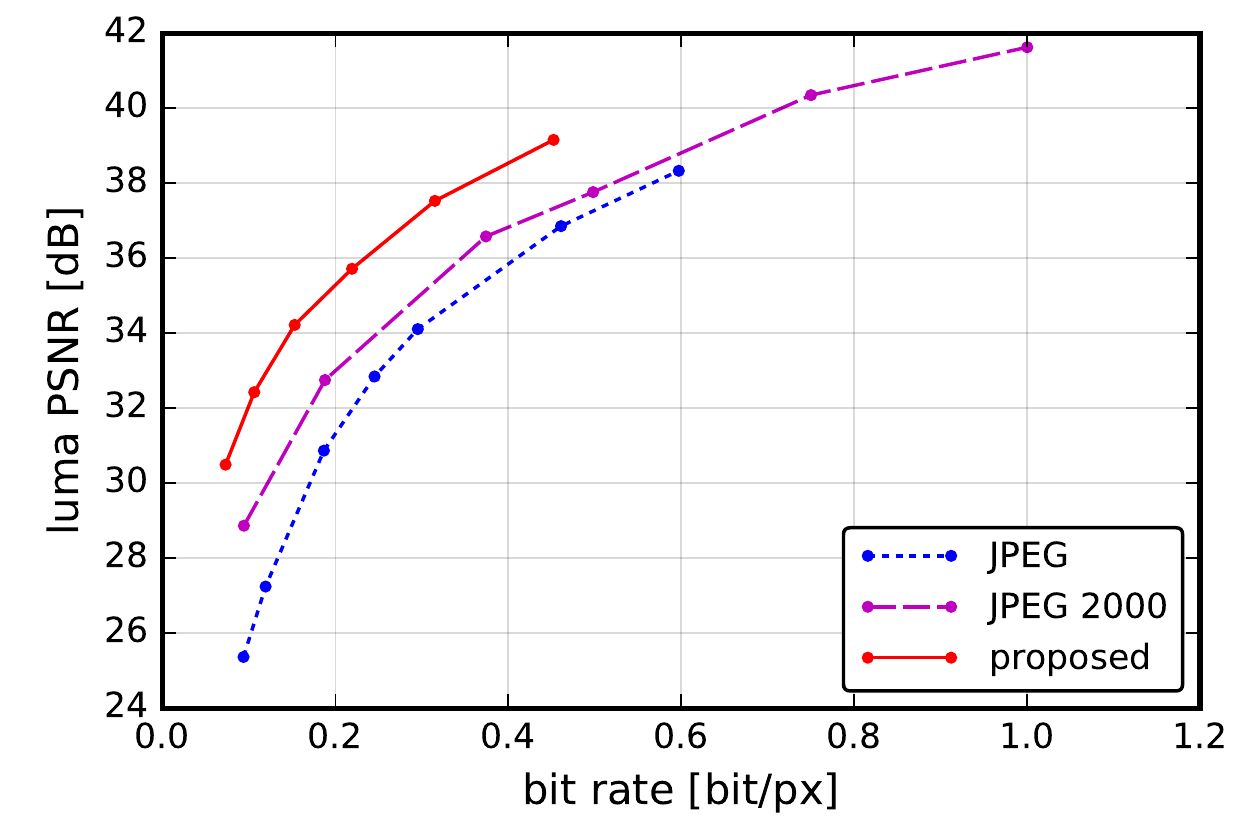}\vspace{1em}\\%
\includegraphics[width=\textwidth]{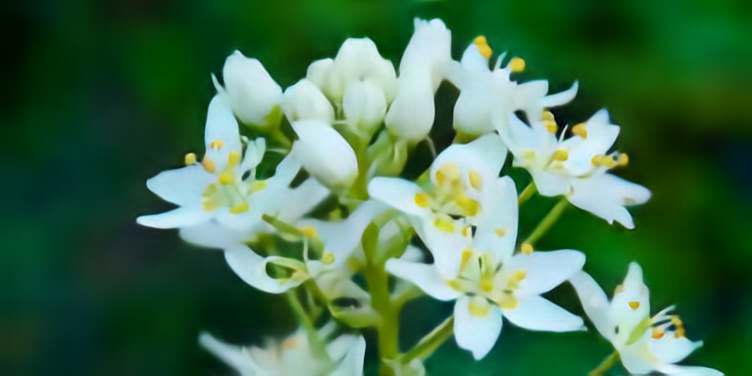}\\%
{\bf Proposed method}, 3749 bytes (0.106 bit/px), PSNR: luma 32.43 dB/chroma 34.00 dB, MS-SSIM: 0.9767\vspace{1em}\\%
\includegraphics[width=\textwidth]{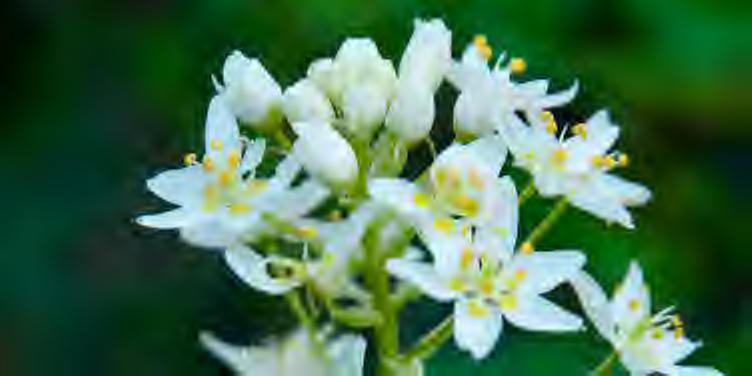}\\%
{\bf JPEG 2000}, 3769 bytes (0.107 bit/px), PSNR: luma 29.49 dB/chroma 32.99 dB, MS-SSIM: 0.9520%
\caption{RGB example, from our personal collection, downsampled and cropped to $752\times 376$ pixels.}
\label{fig:flowers}
\end{figure}

\begin{figure}[p]
\centering\footnotesize%
\includegraphics[width=.5\textwidth]{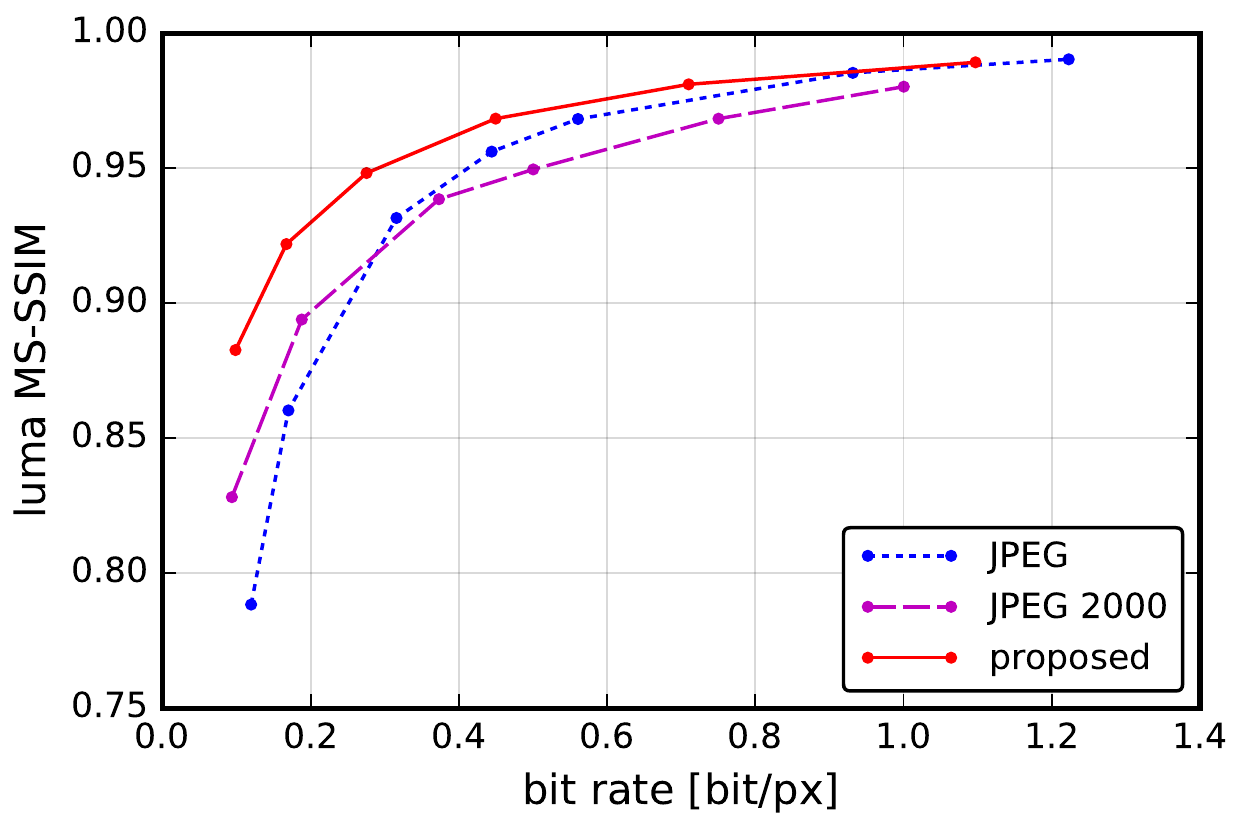}\hfill%
\includegraphics[width=.5\textwidth]{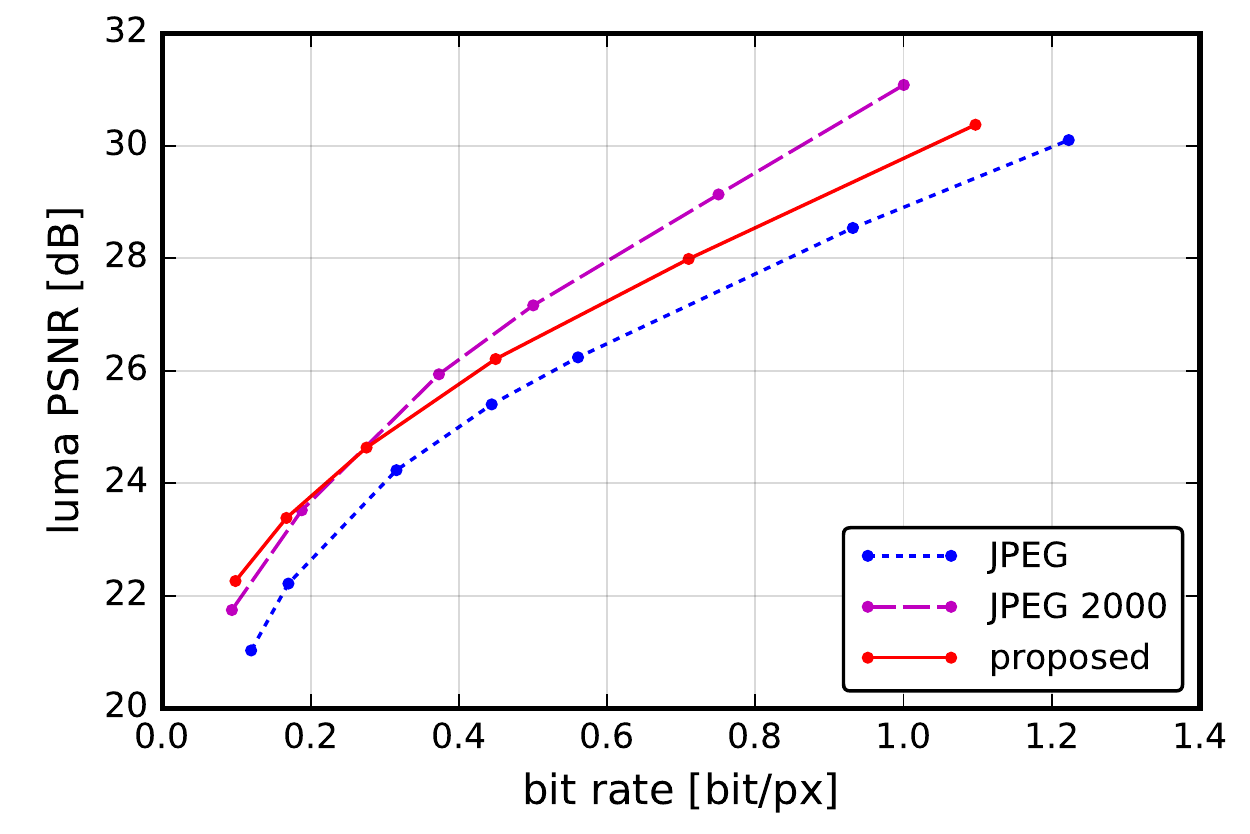}\vspace{1em}\\%
\includegraphics[width=\textwidth]{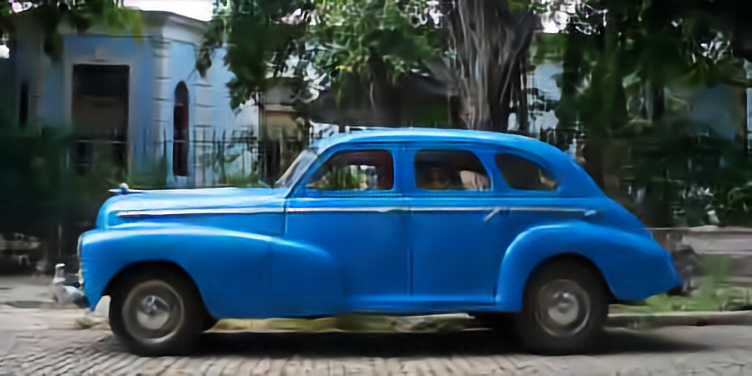}\\%
{\bf Proposed method}, 5908 bytes (0.167 bit/px), PSNR: luma 23.38 dB/chroma 31.86 dB, MS-SSIM: 0.9219\vspace{1em}\\%
\includegraphics[width=\textwidth]{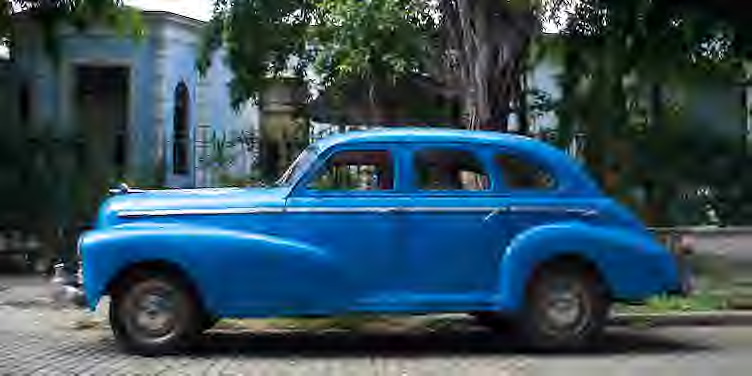}\\%
{\bf JPEG 2000}, 5908 bytes (0.167 bit/px), PSNR: luma 23.24 dB/chroma 31.04 dB, MS-SSIM: 0.8803%
\caption{RGB example, from our personal collection, downsampled and cropped to $752\times 376$ pixels.}
\label{fig:car}
\end{figure}

\begin{figure}[p]
\centering\footnotesize%
\includegraphics[width=.5\textwidth]{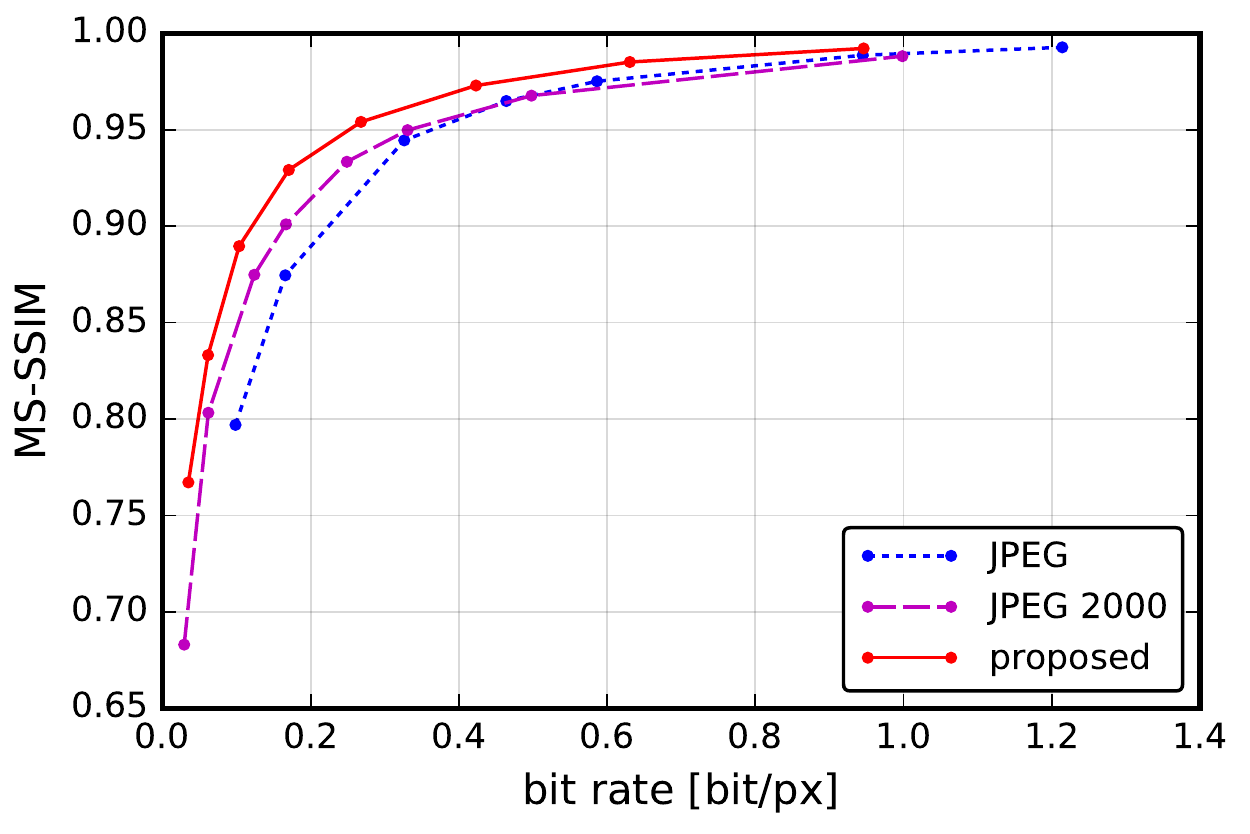}\hfill%
\includegraphics[width=.5\textwidth]{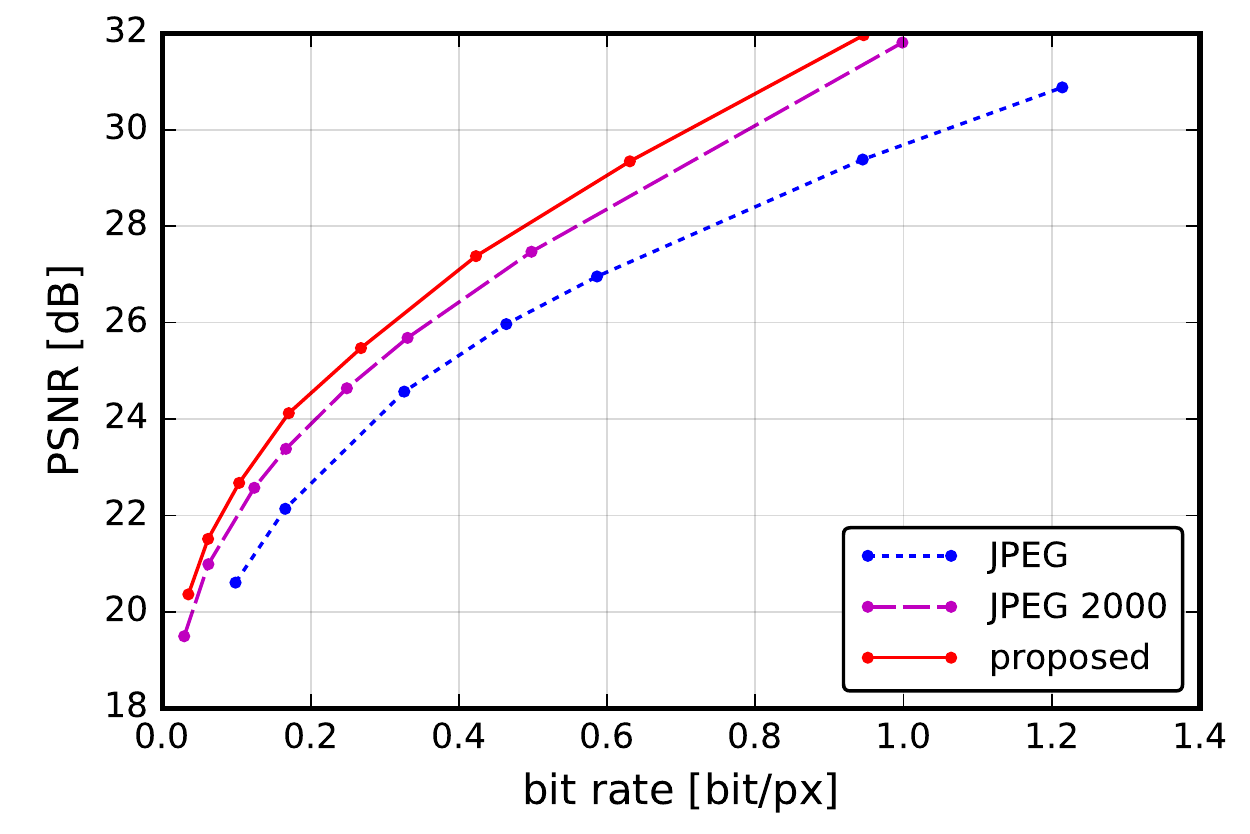}\vspace{1em}\\%
\includegraphics[width=\textwidth]{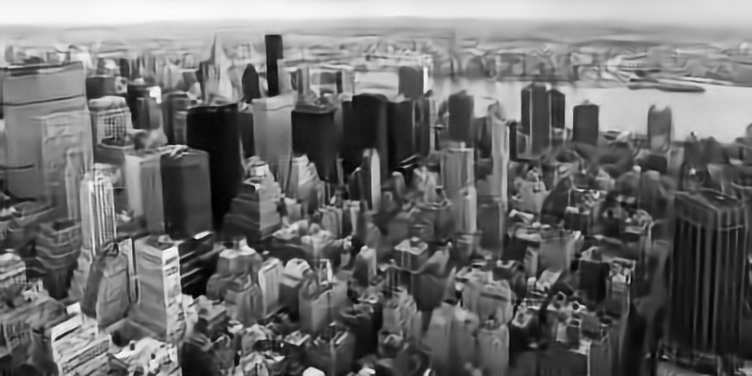}\\%
{\bf Proposed method}, 6021 bytes (0.170 bit/px), PSNR: 24.12 dB, MS-SSIM: 0.9292\vspace{1em}\\%
\includegraphics[width=\textwidth]{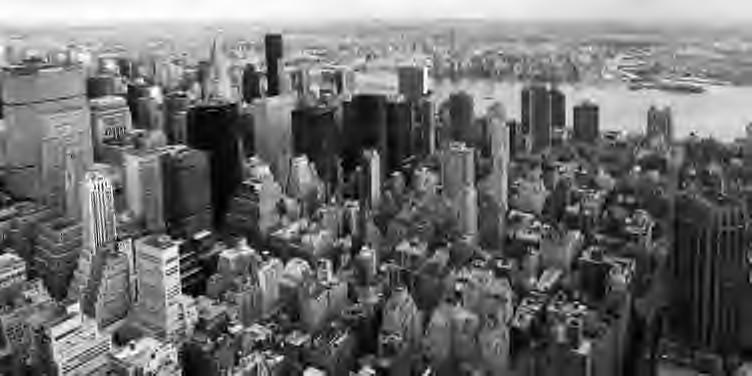}\\%
{\bf JPEG 2000}, 6037 bytes (0.171 bit/px), PSNR: 23.47 dB, MS-SSIM: 0.9036%
\caption{Grayscale example, from our personal collection, downsampled and cropped to $752\times 376$ pixels.}
\label{fig:nyc}
\end{figure}

Although it is desirable to summarize and compare the rate--distortion behavior of JPEG, JPEG~2000, and our method across an image set, it is difficult to do this in a way that is fair and interpretable. First, rate--distortion behavior varies substantially across bit rates for different images. For example, for the image in figure~\ref{fig:flowers}, our method achieves the same MSE with roughly 50\% of the bits needed by JPEG~2000 for low rates, and about 30\% for high rates. For the image in figure~\ref{fig:nyc}, the gains are more modest, although still significant through the range. But for the image in figure~\ref{fig:car}, our method only slightly outperforms JPEG~2000 at low rates, and under-performs it at high rates. Note that these behaviors are again different for MS-SSIM, which shows a significant improvement for all images and bit rates (consistent with their visual appearance).

\begin{figure}[t]
\centering%
\includegraphics[width=\textwidth,trim=0 10 0 10]{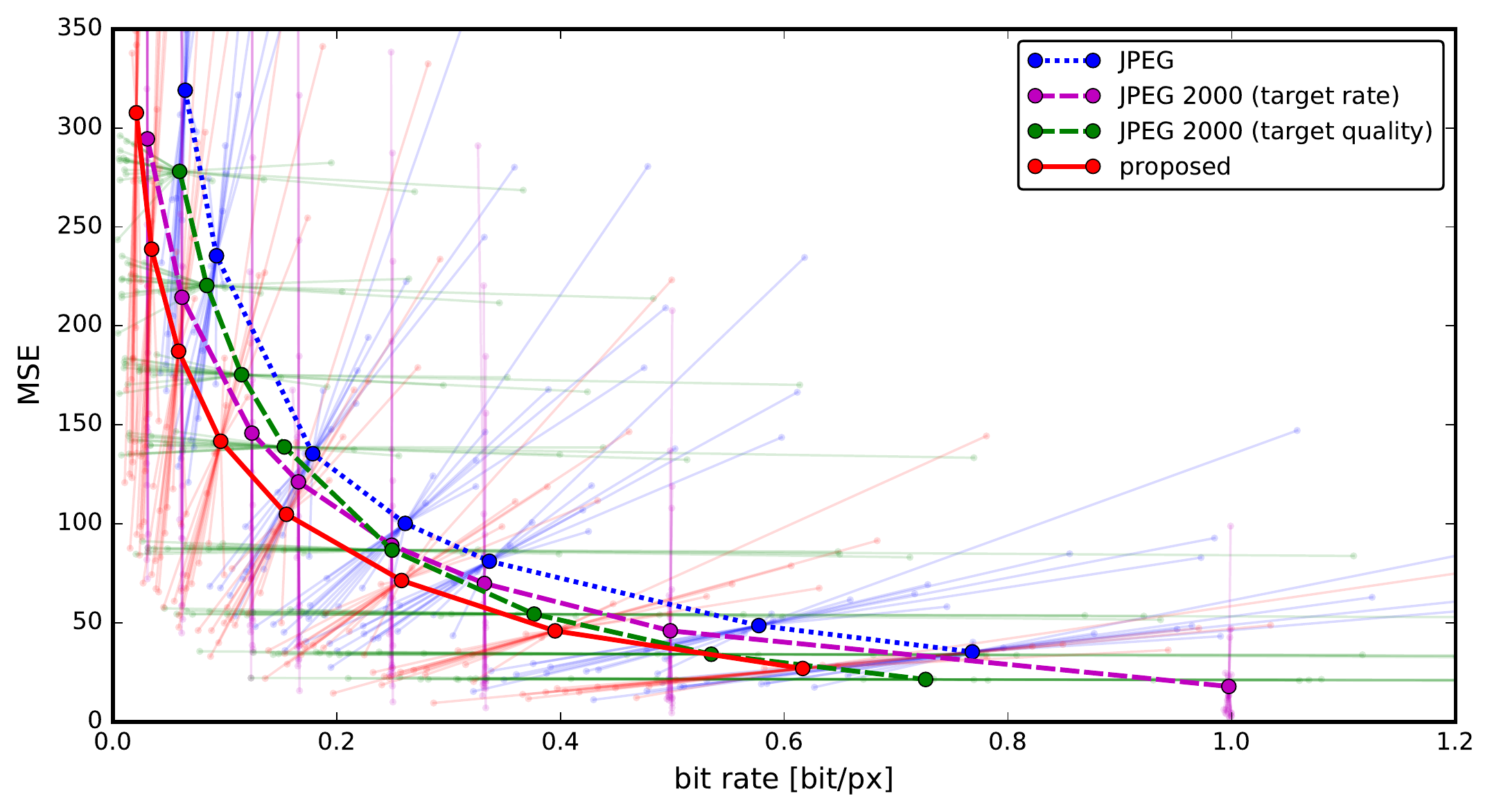}
\caption{Summary rate--distortion curves, computed by averaging results over the 24 images in the Kodak test set. Each point is connected by translucent lines to the set of 24 points corresponding to the individual image R-D values from which it was derived. JPEG results are averaged over images compressed with identical quality settings. Results of the proposed method are averaged over images compressed with identical $\lambda$ values (and thus, computed with exactly the same forward and inverse transforms). The two JPEG~2000 curves are computed with the same implementation, by averaging over images compressed with the same target rate or the same target quality. Note that these two methods of selecting points to be averaged lead to significantly different average results.}
\label{fig:kodak_scatter}
\end{figure}

Second, there is no obvious or agreed-upon method for combining rate--distortion curves across images. More specifically, one must decide which points in the curves to combine. For our method, it is natural to average the MSE and entropy values across images compressed using the same choice of $\lambda$, since these are all coded and decoded using exactly the same representation and quantization scheme. For JPEG, it seems natural to average over images coded at the same “quality” setting, which appear to be coded using the same quantization choices. The OpenJPEG implementation of JPEG~2000 we use allows selection of points on the rate--distortion curve \emph{either} through specification of a target bit rate, \emph{or} a target quality. This choice has no effect on rate--distortion plots for individual images (verified, but not shown), but has a substantial effect when averaging over images, since the two choices lead one to average over a different set of R-D points. This is illustrated in figure~\ref{fig:kodak_scatter}. Even if points were selected in exactly the same fashion for each of the methods (say, matched to a given set of target rates), summary plots can still over- or underemphasize high rate vs. low rate performance.

We conclude that summaries of rate--distortion are of limited use. Instead, we encourage the reader to browse our extensive collection of test images, with individual rate--distortion plots for each image, available at \url{http://www.cns.nyu.edu/~lcv/iclr2017} in both grayscale and RGB.

In the following pages, we show additional example images, compressed at relatively low bit rates, in order to visualize the qualitative nature of compression artifacts. On each page, the JPEG~2000 image is selected to have the lowest possible bit rate that is equal or greater than the bit rate of the proposed method. In all experiments, we compare to JPEG with 4:2:0 chroma subsampling, and the OpenJPEG implementation of JPEG~2000 with the default “multiple component transform”. For evaluating PSNR, we use the JPEG-defined conversion matrix to convert between RGB and Y'CbCr. For evaluating MS-SSIM~\citep{WaSiBo03}, we used only the resulting luma component. Original images are not shown, but are available online, along with compressed images at a variety of other bit rates, at \url{http://www.cns.nyu.edu/\~lcv/iclr2017}.

\begin{figure}[p]
\centering\footnotesize%
\includegraphics[width=.5\textwidth]{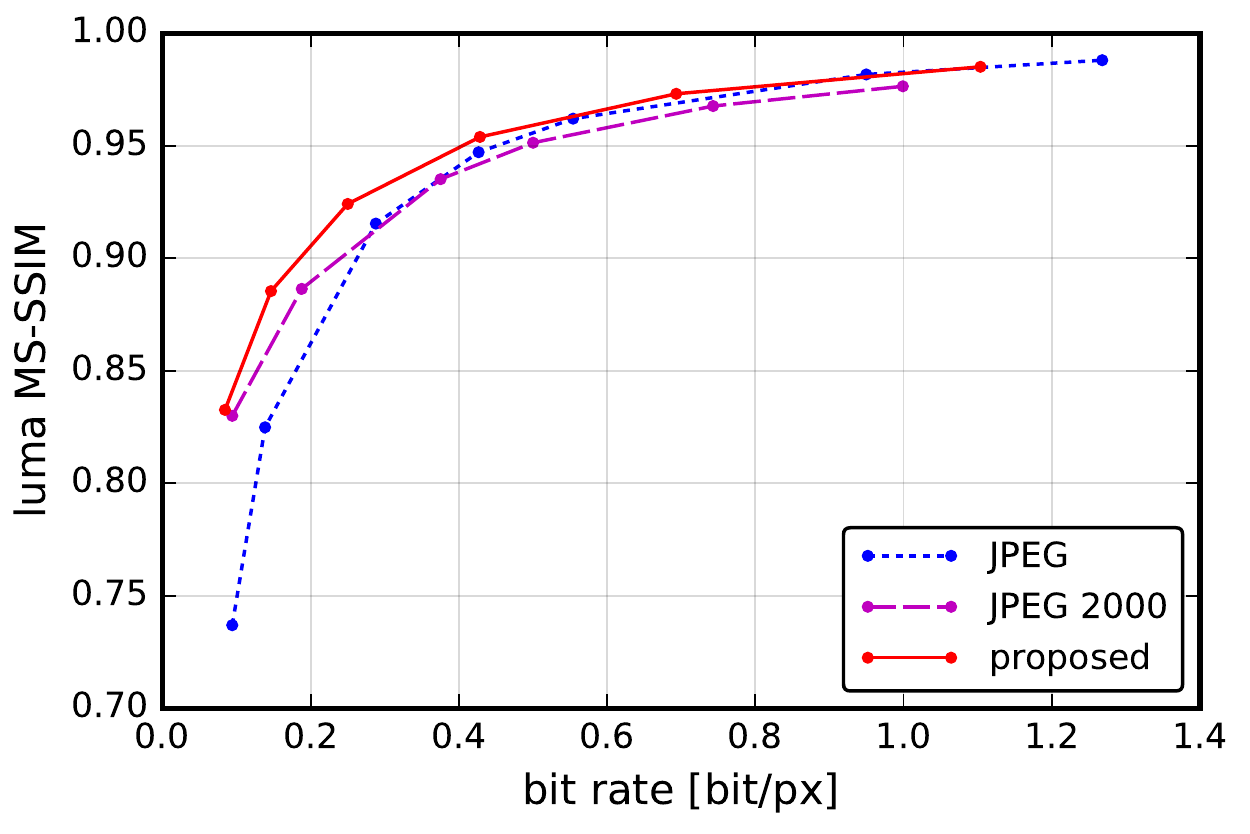}\hfill%
\includegraphics[width=.5\textwidth]{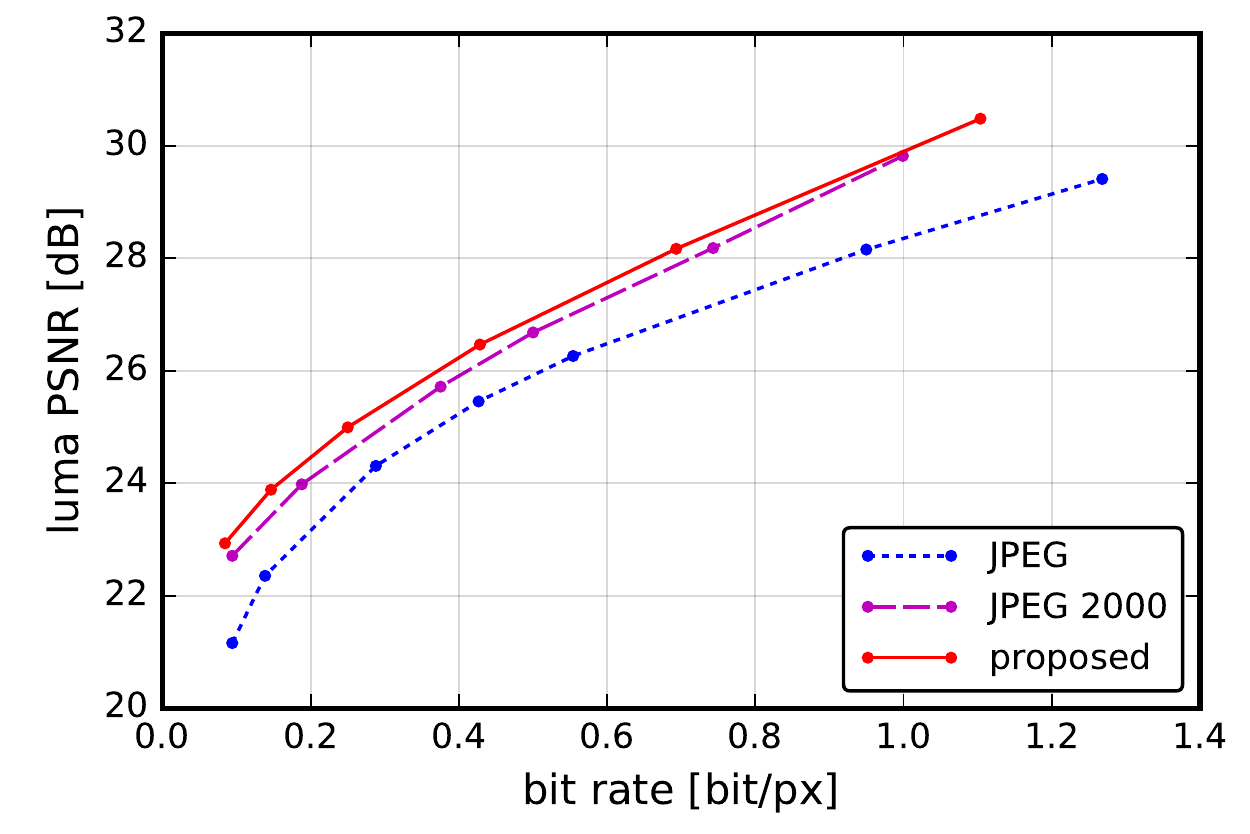}\vspace{1em}\\%
\includegraphics[width=\textwidth]{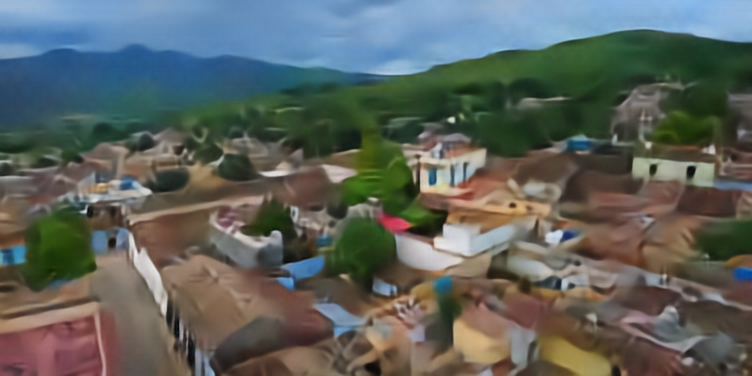}\\%
{\bf Proposed method}, 2978 bytes (0.084 bit/px), PSNR: luma 22.93 dB/chroma 31.45 dB, MS-SSIM: 0.8326\vspace{1em}\\%
\includegraphics[width=\textwidth]{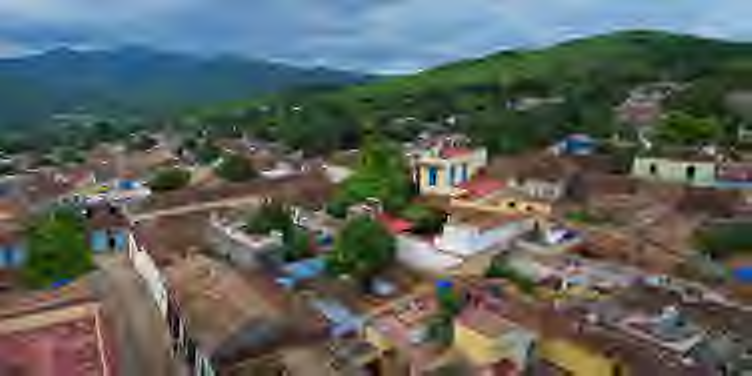}\\%
{\bf JPEG 2000}, 2980 bytes (0.084 bit/px), PSNR: luma 22.53 dB/chroma 31.09 dB, MS-SSIM: 0.8225%
\caption{RGB example, from our personal collection, downsampled and cropped to $752\times 376$ pixels.}
\end{figure}

\begin{figure}[p]
\centering\footnotesize%
\includegraphics[width=.5\textwidth]{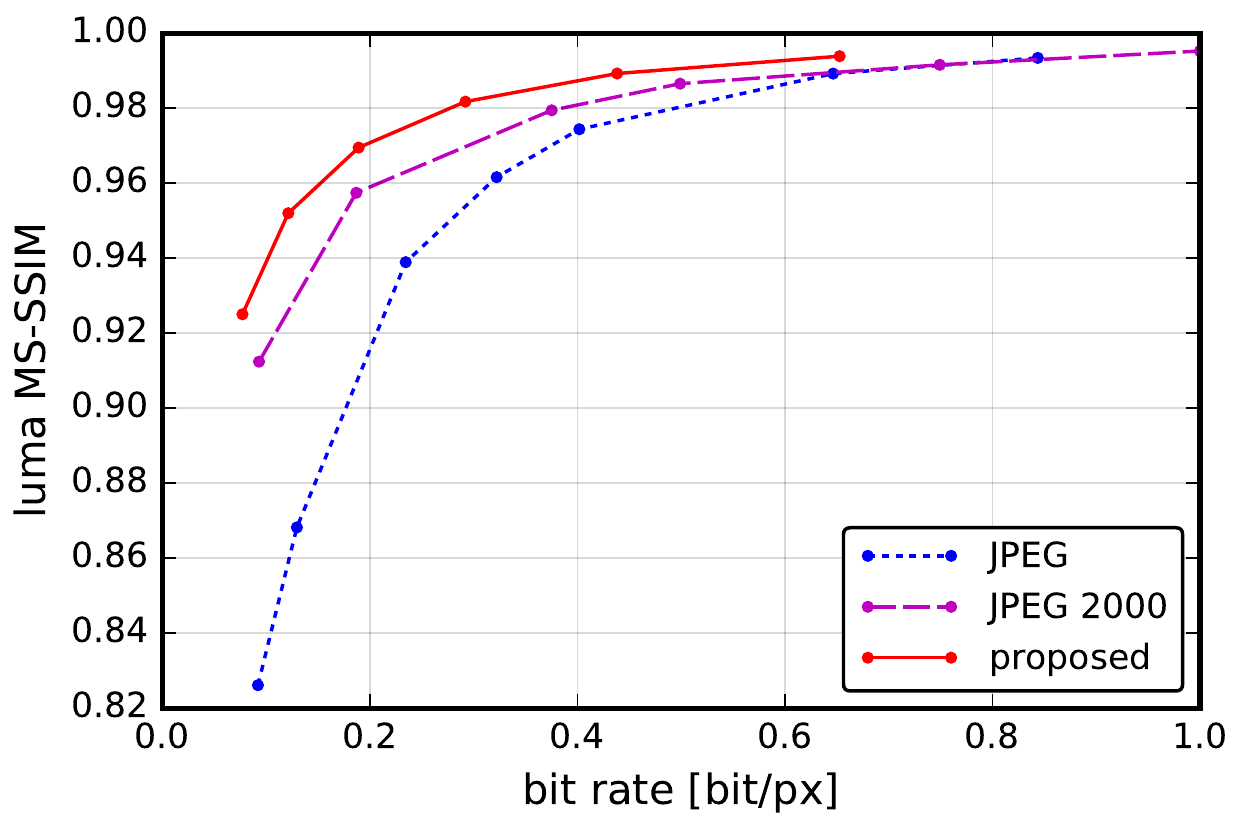}\hfill%
\includegraphics[width=.5\textwidth]{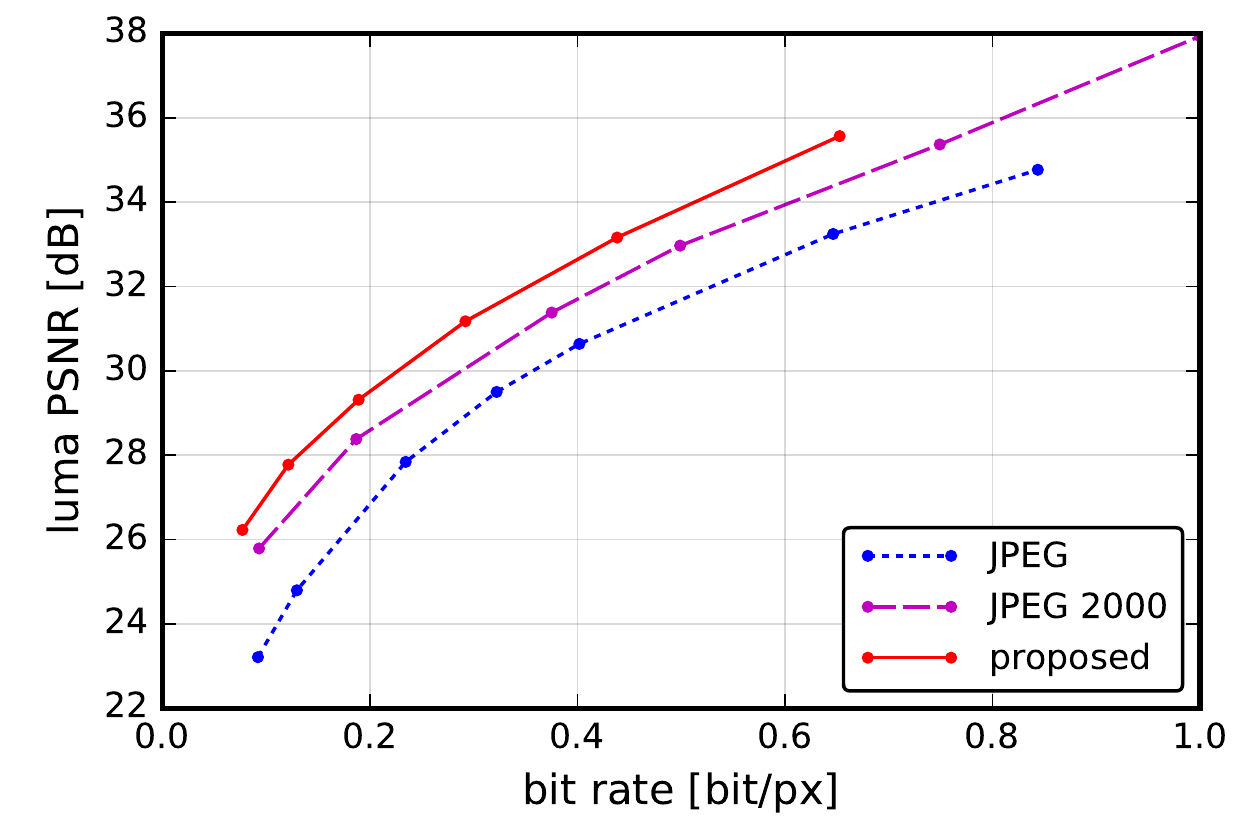}\vspace{1em}\\%
\includegraphics[width=\textwidth,height=.3\textheight,keepaspectratio]{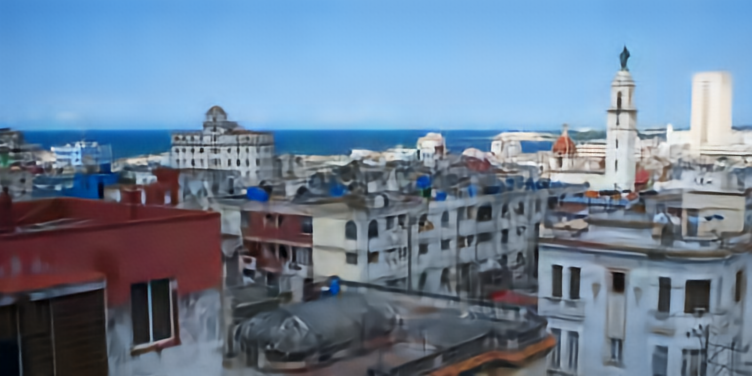}\\%
{\bf Proposed method}, 6680 bytes (0.189 bit/px), PSNR: luma 29.31 dB/chroma 36.17 dB, MS-SSIM: 0.9695\vspace{1em}\\%
\includegraphics[width=\textwidth,height=.3\textheight,keepaspectratio]{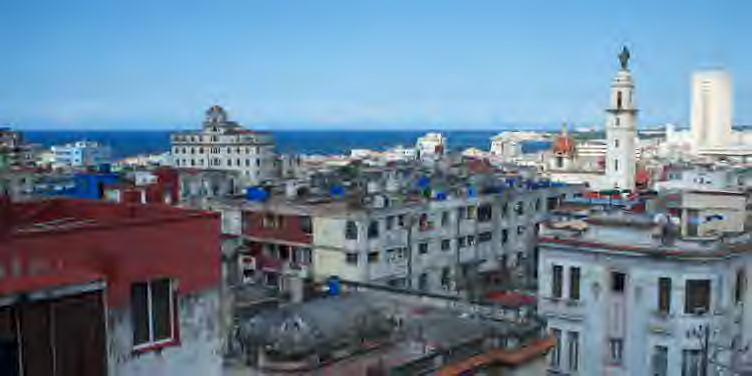}\\%
{\bf JPEG 2000}, 6691 bytes (0.189 bit/px), PSNR: luma 28.45 dB/chroma 35.32 dB, MS-SSIM: 0.9586%
\caption{RGB example, from our personal collection, downsampled and cropped to $752\times 376$ pixels.}
\end{figure}

\begin{figure}[p]
\centering\footnotesize%
\includegraphics[width=.5\textwidth]{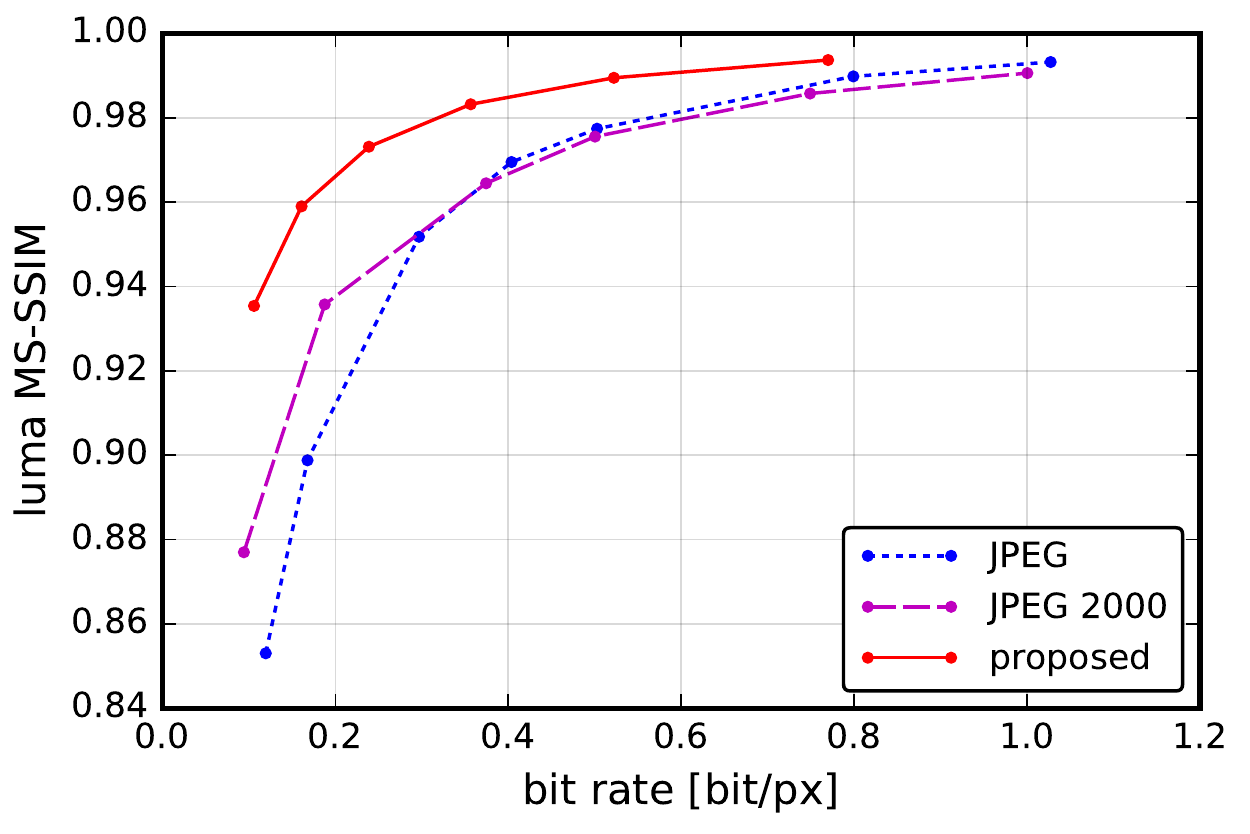}\hfill%
\includegraphics[width=.5\textwidth]{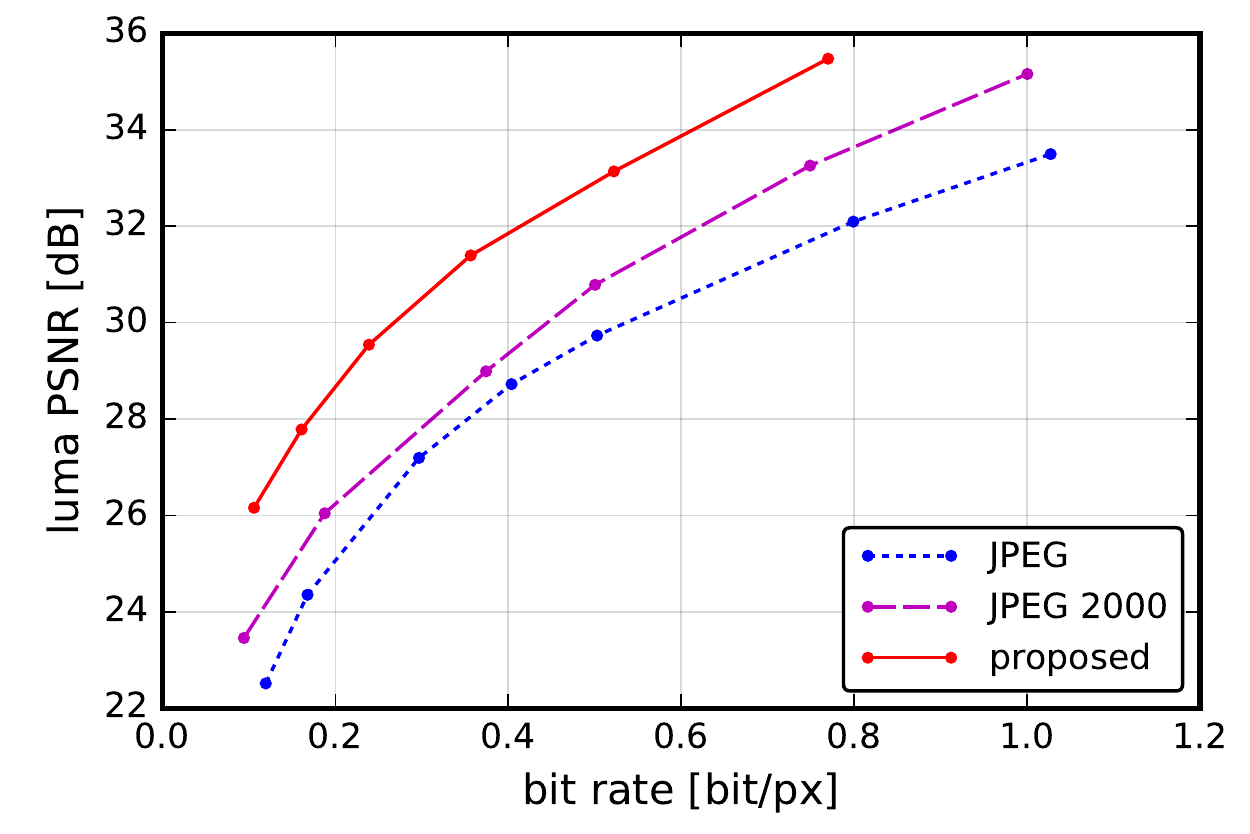}\vspace{1em}\\%
\includegraphics[width=\textwidth]{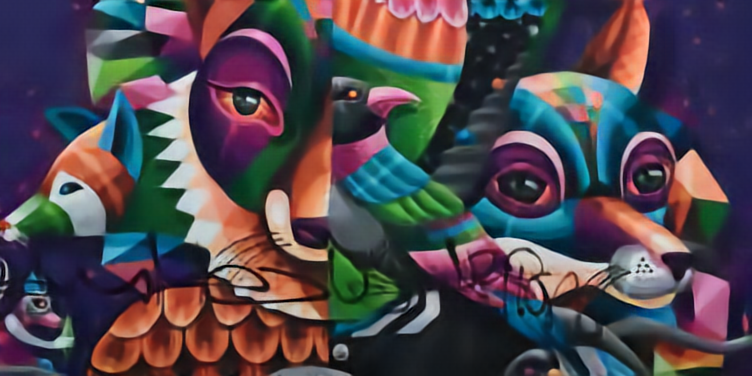}\\%
{\bf Proposed method}, 5683 bytes (0.161 bit/px), PSNR: luma 27.78 dB/chroma 32.60 dB, MS-SSIM: 0.9590\vspace{1em}\\%
\includegraphics[width=\textwidth]{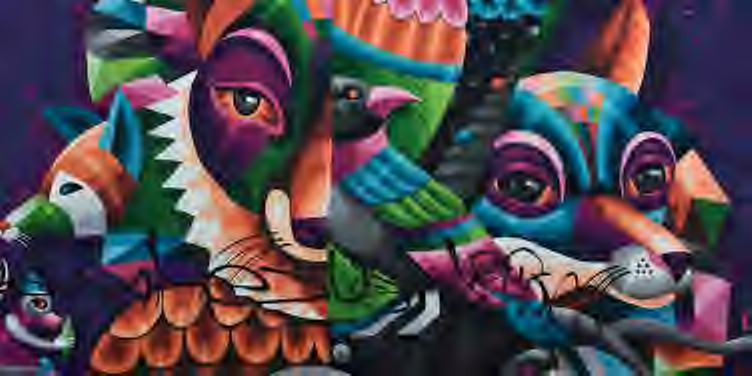}\\%
{\bf JPEG 2000}, 5724 bytes (0.162 bit/px), PSNR: luma 25.36 dB/chroma 31.20 dB, MS-SSIM: 0.9202%
\caption{RGB example, from our personal collection, downsampled and cropped to $752\times 376$ pixels.}
\end{figure}

\begin{figure}[p]
\centering\footnotesize%
\includegraphics[width=.5\textwidth]{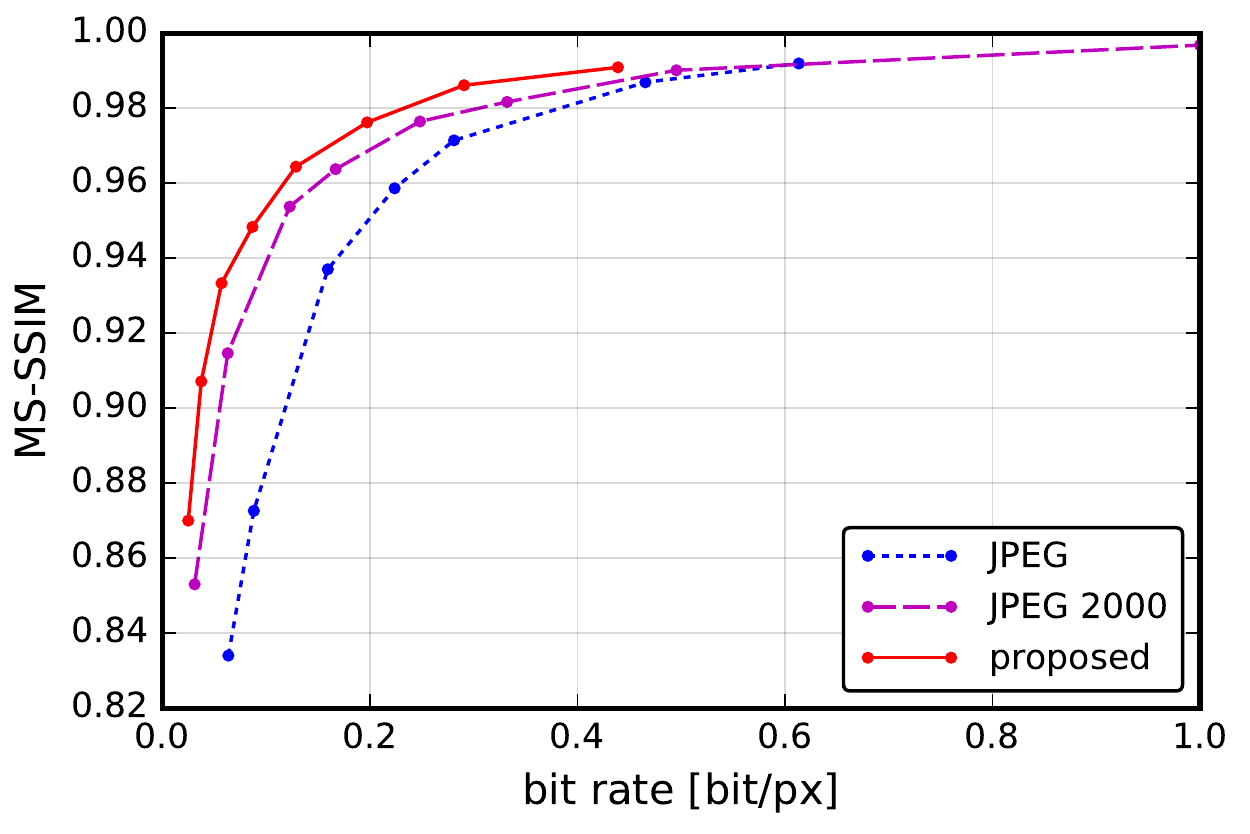}\hfill%
\includegraphics[width=.5\textwidth]{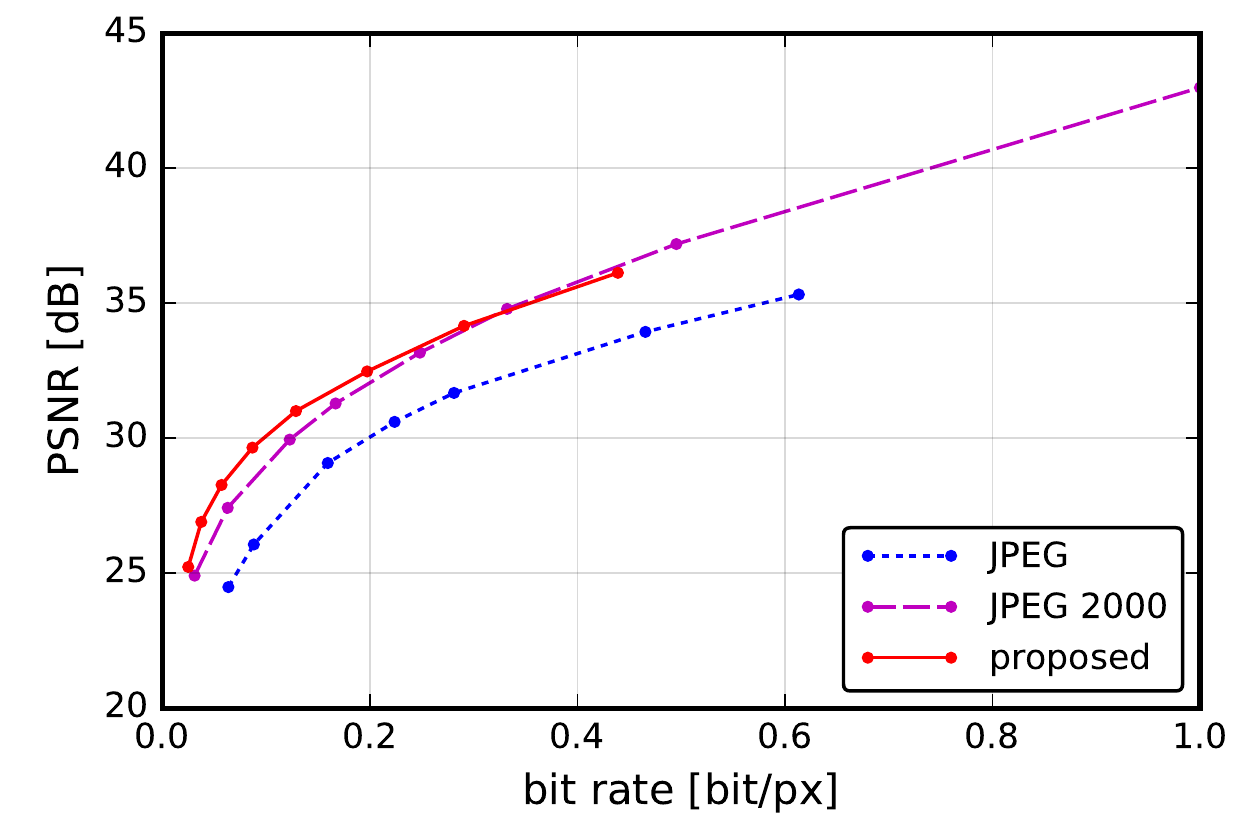}\vspace{1em}\\%
\includegraphics[width=\textwidth]{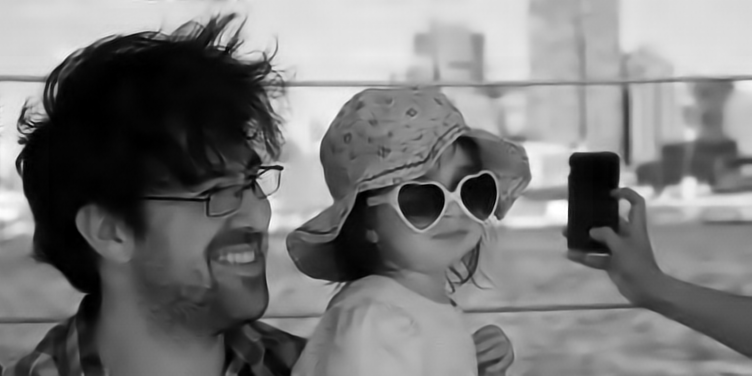}\\%
{\bf Proposed method}, 4544 bytes (0.129 bit/px), PSNR: 31.01 dB, MS-SSIM: 0.9644\vspace{1em}\\%
\includegraphics[width=\textwidth]{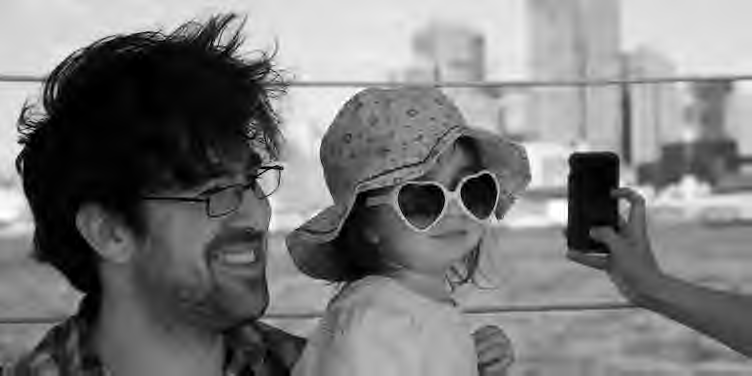}\\%
{\bf JPEG 2000}, 4554 bytes (0.129 bit/px), PSNR: 30.17 dB, MS-SSIM: 0.9546%
\caption{Grayscale example, from our personal collection, downsampled and cropped to $752\times 376$ pixels.}
\end{figure}

\begin{figure}[p]
\centering\footnotesize%
\includegraphics[width=.5\textwidth]{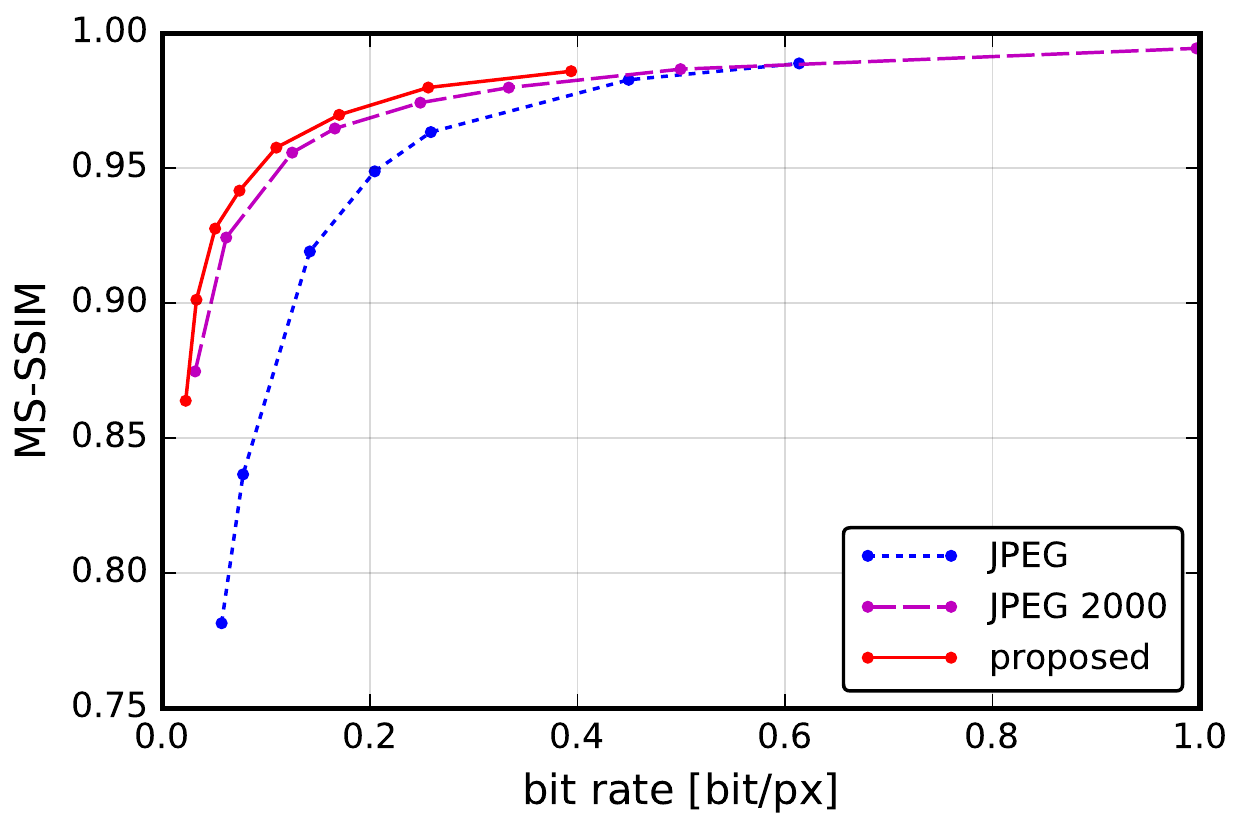}\hfill%
\includegraphics[width=.5\textwidth]{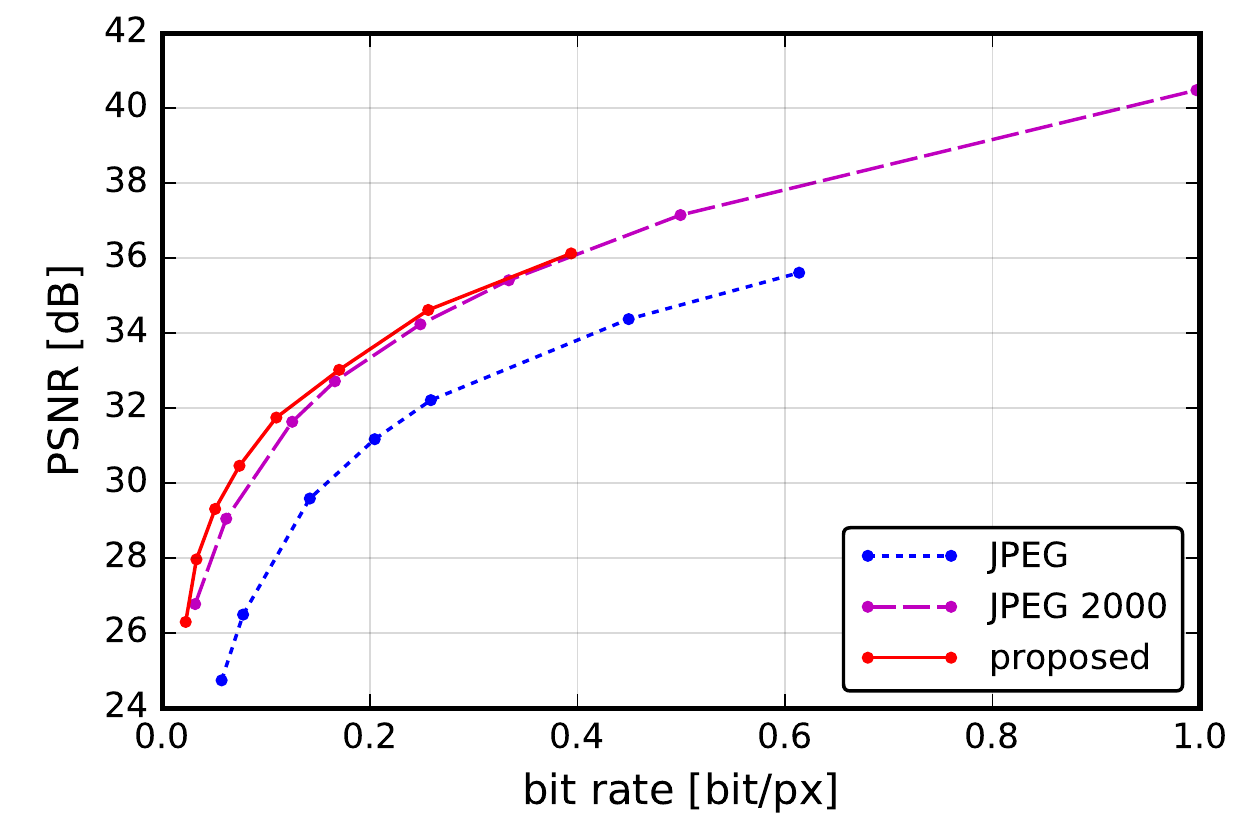}\vspace{1em}\\%
\includegraphics[width=\textwidth]{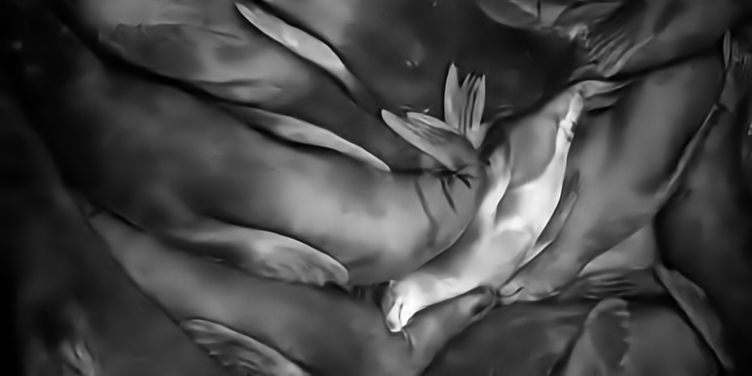}\\%
{\bf Proposed method}, 3875 bytes (0.110 bit/px), PSNR: 31.75 dB, MS-SSIM: 0.9577\vspace{1em}\\%
\includegraphics[width=\textwidth]{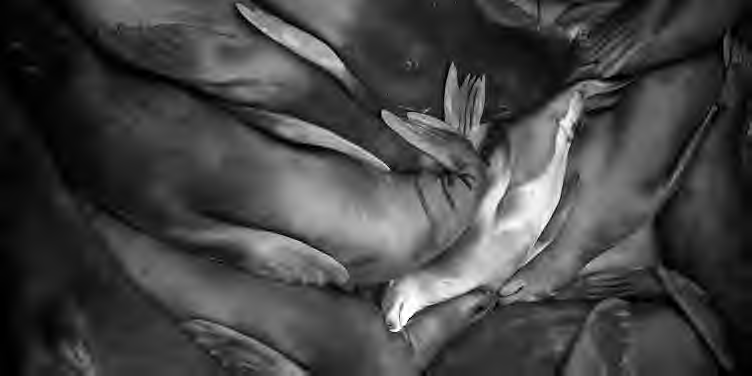}\\%
{\bf JPEG 2000}, 3877 bytes (0.110 bit/px), PSNR: 31.24 dB, MS-SSIM: 0.9511%
\caption{Grayscale example, from our personal collection, downsampled and cropped to $752\times 376$ pixels.}
\end{figure}

\begin{figure}[p]
\centering\footnotesize%
\includegraphics[width=.5\textwidth]{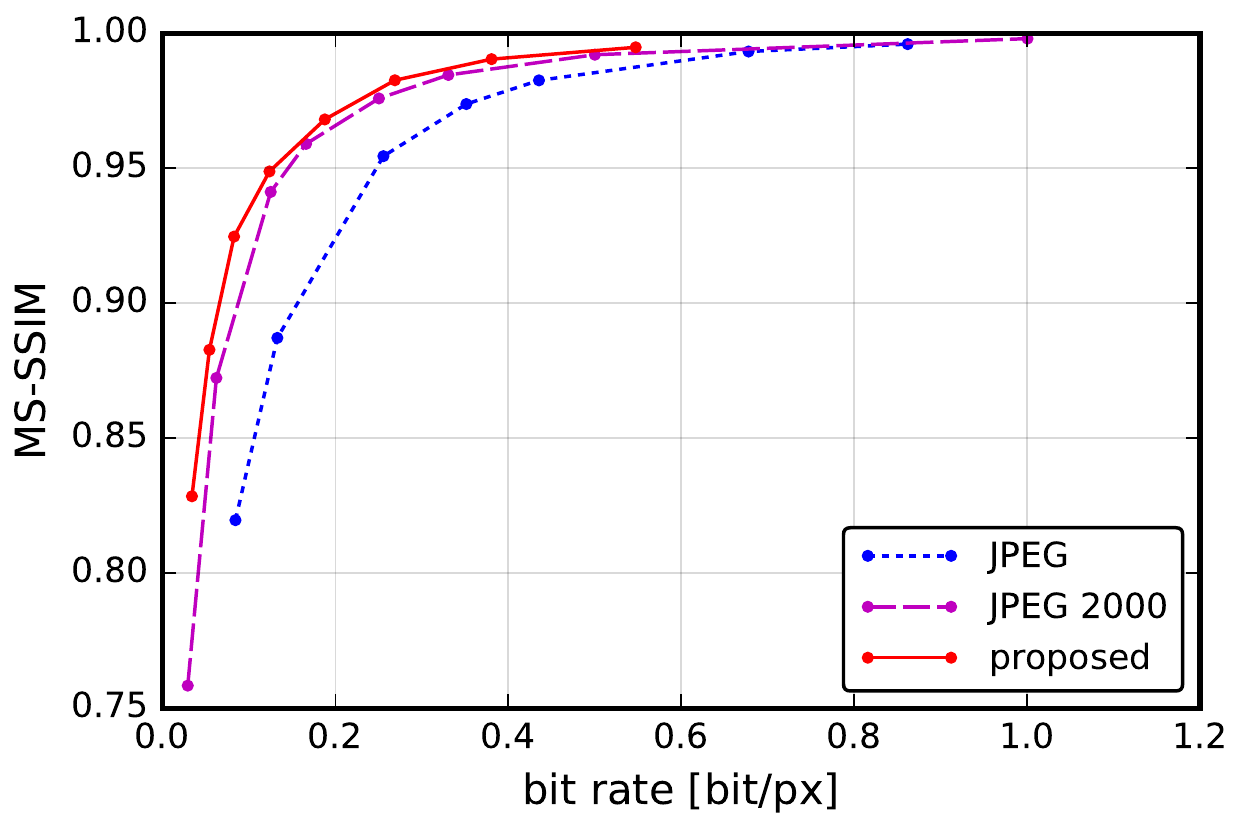}\hfill%
\includegraphics[width=.5\textwidth]{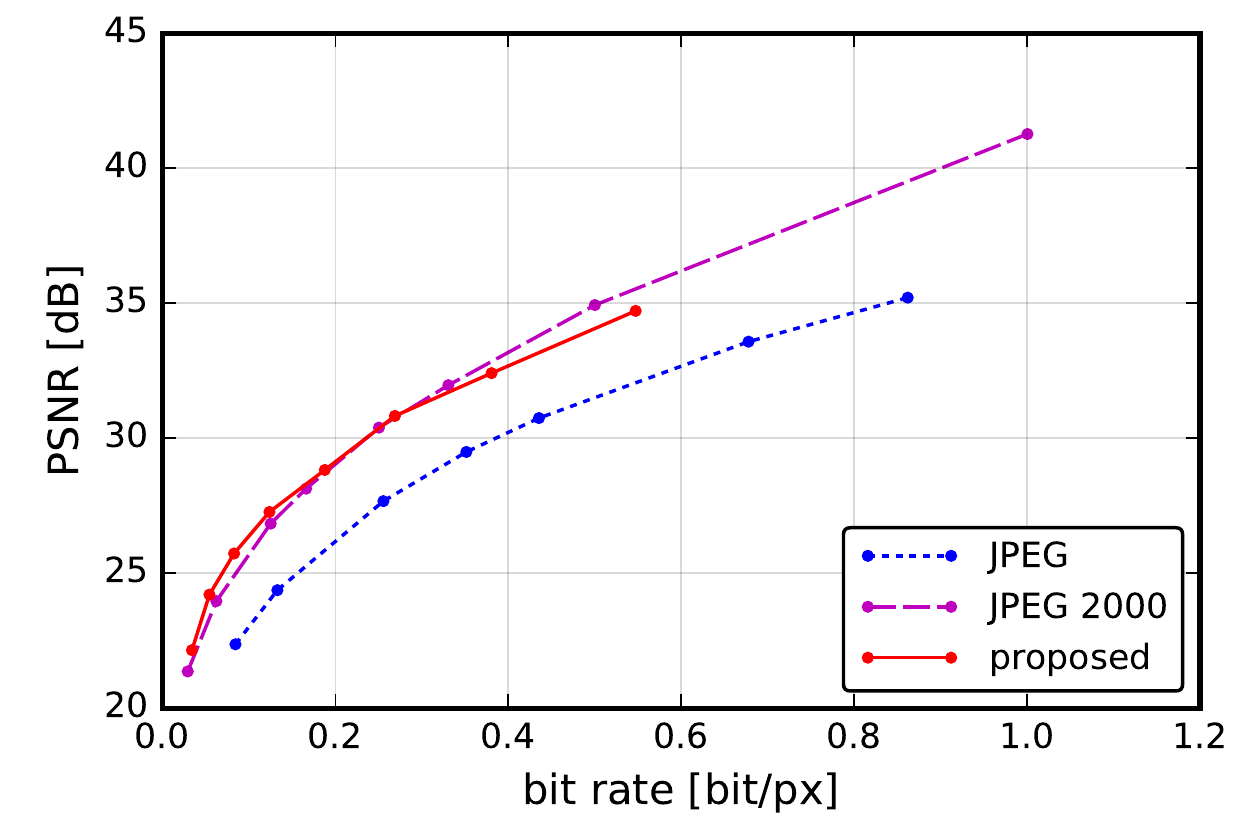}\vspace{1em}\\%
\includegraphics[width=\textwidth]{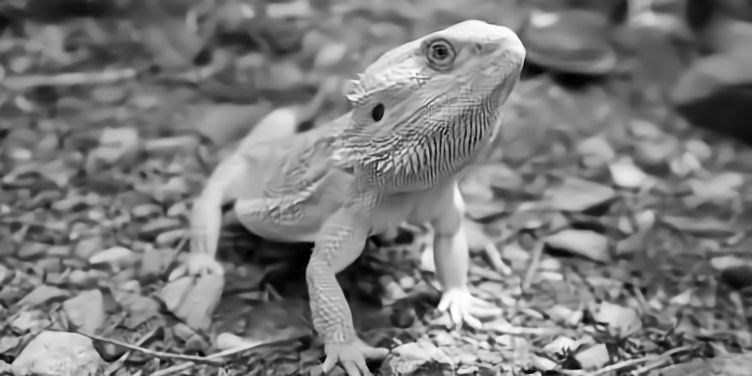}\\%
{\bf Proposed method}, 6633 bytes (0.188 bit/px), PSNR: 28.83 dB, MS-SSIM: 0.9681\vspace{1em}\\%
\includegraphics[width=\textwidth]{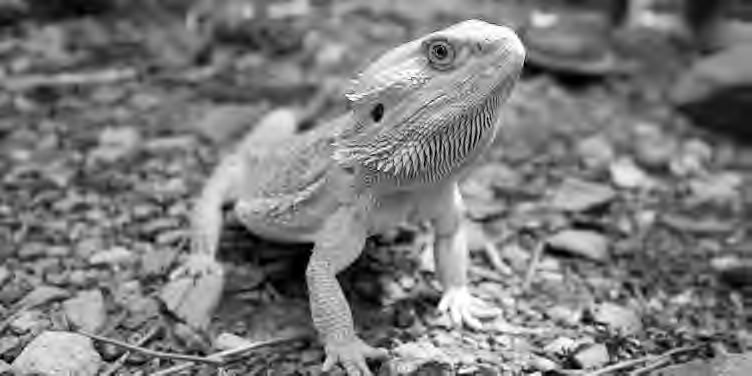}\\%
{\bf JPEG 2000}, 6691 bytes (0.189 bit/px), PSNR: 28.83 dB, MS-SSIM: 0.9651%
\caption{Grayscale example, from our personal collection, downsampled and cropped to $752\times 376$ pixels.}
\label{fig:lizard}
\end{figure}

\begin{figure}[p]
\centering\footnotesize%
\includegraphics[width=.5\textwidth]{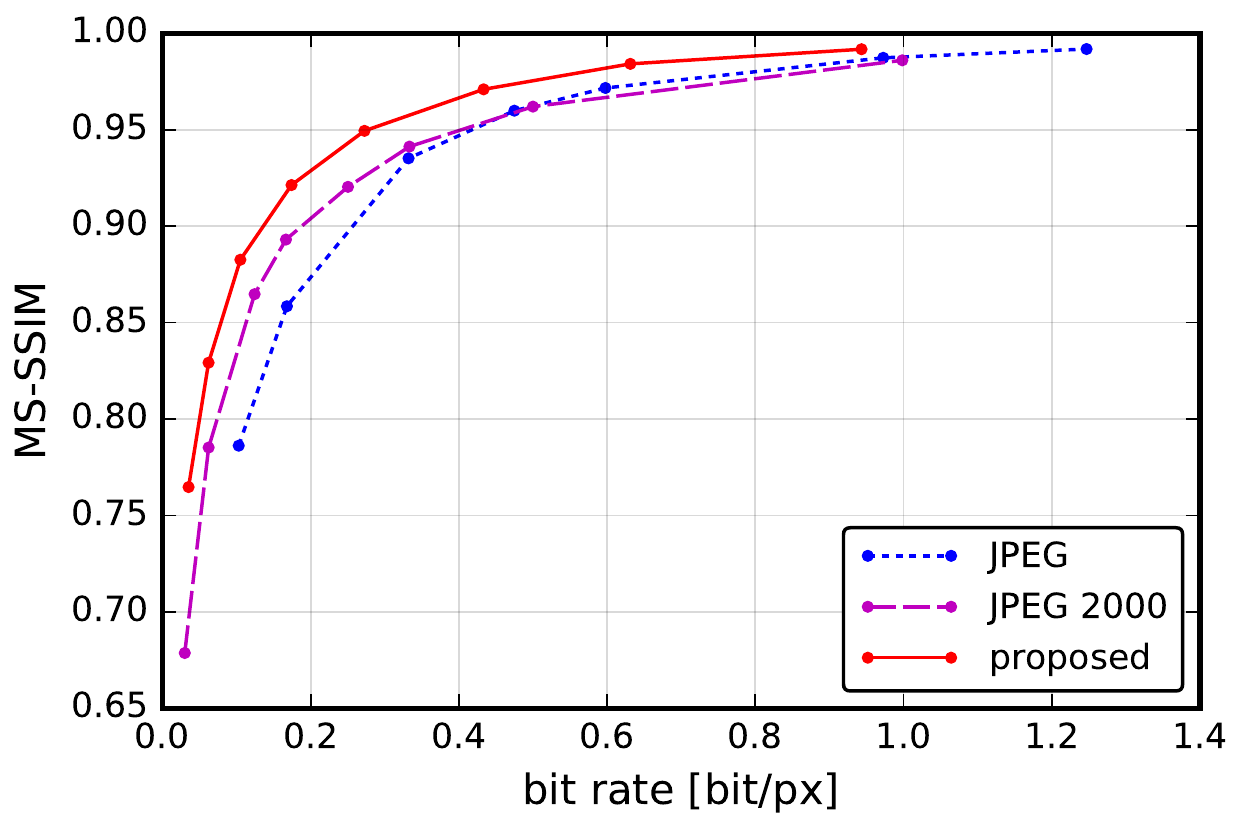}\hfill%
\includegraphics[width=.5\textwidth]{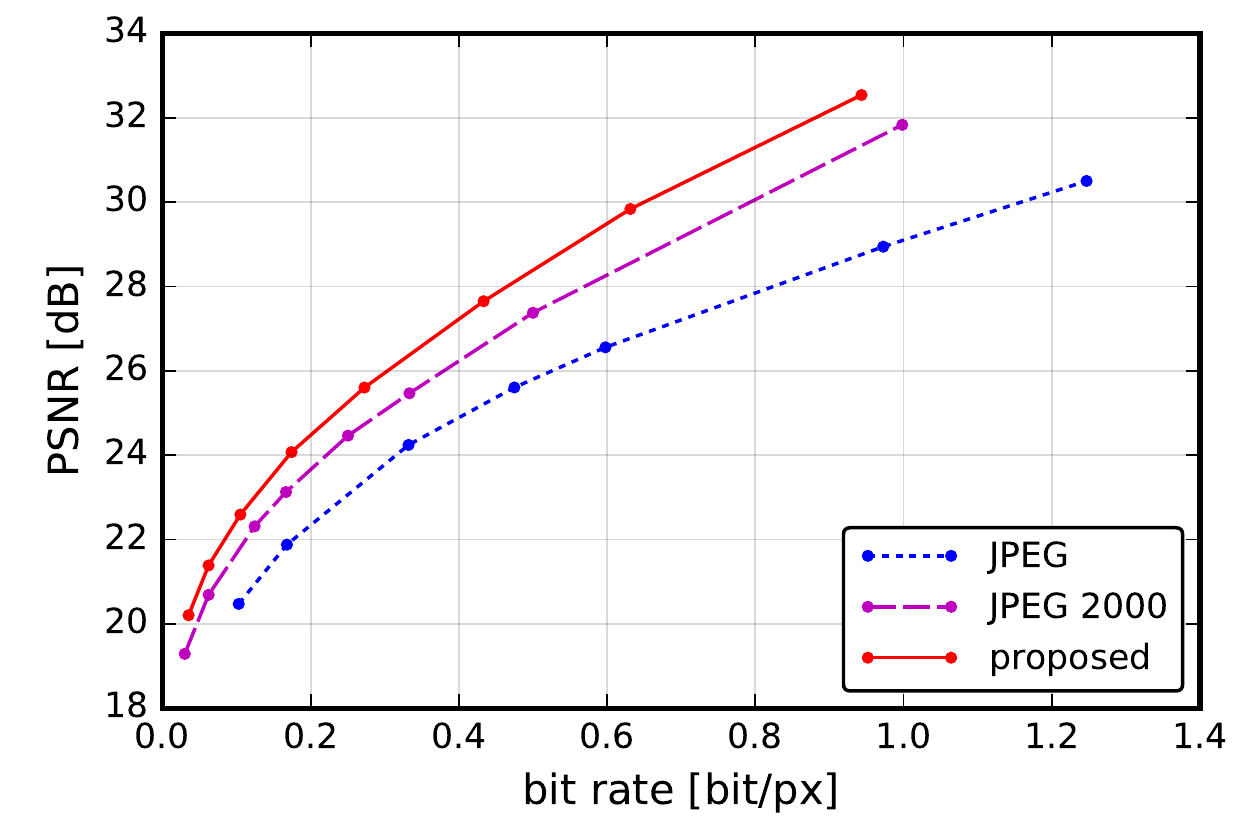}\vspace{1em}\\%
\includegraphics[width=\textwidth]{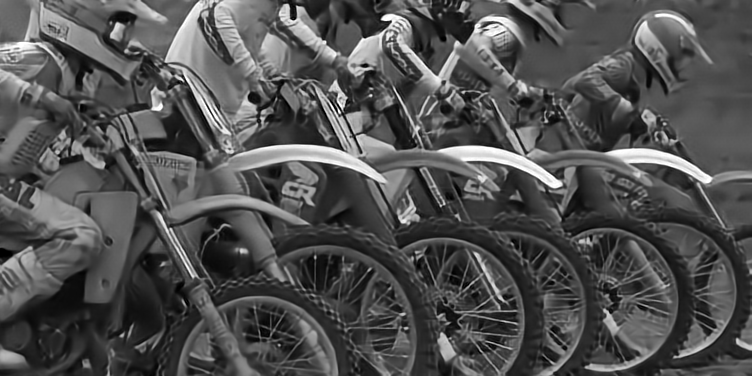}\\%
{\bf Proposed method}, 10130 bytes (0.287 bit/px), PSNR: 25.27 dB, MS-SSIM: 0.9537\vspace{1em}\\%
\includegraphics[width=\textwidth]{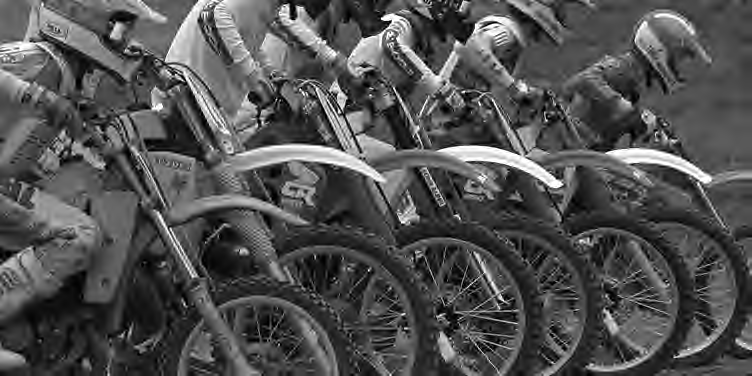}\\%
{\bf JPEG 2000}, 10197 bytes (0.289 bit/px), PSNR: 24.41 dB, MS-SSIM: 0.9320%
\caption{Grayscale example, from the Kodak test set, downsampled and cropped to $752\times 376$ pixels.}
\end{figure}

\end{document}